\documentclass[10pt,twocolumn,letterpaper]{article}

\usepackage[pagenumbers]{cvpr} %

\usepackage{cuted}
\usepackage{caption}
\usepackage{multirow}
\usepackage{subcaption}
\usepackage{makecell}
\usepackage{pgfplots}\pgfplotsset{compat=1.17}
\usepackage[table]{xcolor} 
\usepackage[most]{tcolorbox}

\definecolor{lightgreen}{RGB}{240,248,236}

\definecolor{findingblue}{RGB}{55, 105, 165}

\newtcolorbox[auto counter]{findingbox}[1][]{
    enhanced,
    colback=findingblue!7,
    colframe=findingblue,
    boxrule=1pt,
    arc=2mm,
    left=6pt,
    right=6pt,
    top=2pt,
    bottom=2pt,
    before upper={
        \textbf{\textit{Finding~\thetcbcounter.}}\enspace
    },
    #1
}

\usepackage{xspace}
\newcommand{\ourmodel}{Hi3R-VLA\xspace}

\definecolor{cvprblue}{rgb}{0.21,0.49,0.74}
\usepackage[pagebackref,breaklinks,colorlinks,allcolors=cvprblue]{hyperref}

\def\confName{CVPR}
\def\confYear{2027}

\title{
\ourmodel: Grounding Semantics and Actions in 3D
}

\author{Shizhe Chen, Paul Pacaud, Cordelia Schmid\\
Inria, \'Ecole normale sup\'erieure, CNRS, PSL Research University\\
\small {\url{https://cshizhe.github.io/projects/hi3r_vla.html}}
}

\begin{document}
\maketitle

\begin{strip}
    \centering
    \includegraphics[width=\linewidth]{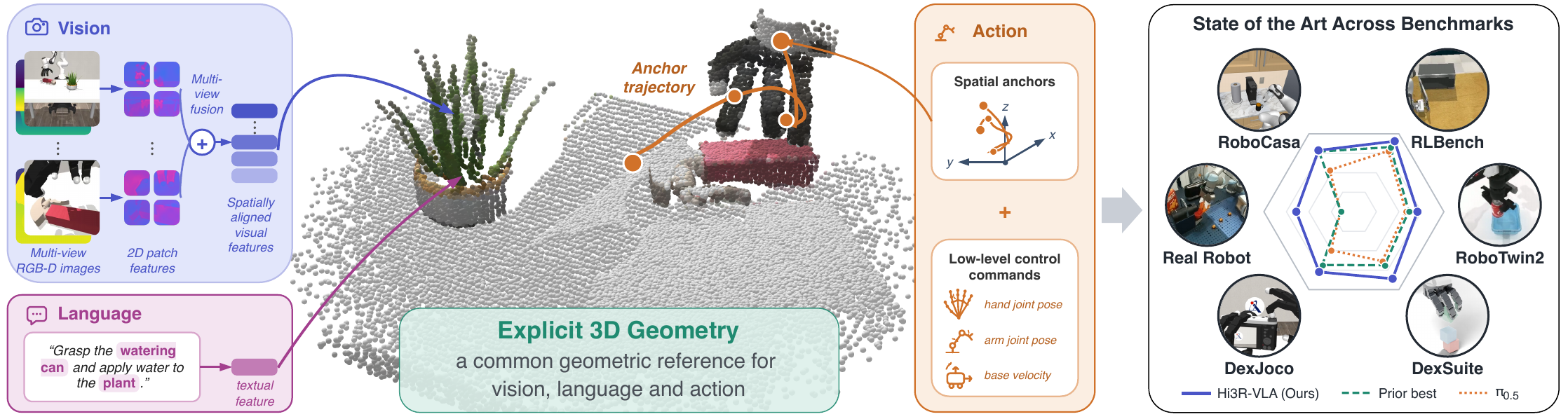}
    \captionsetup{hypcap=false}
    \captionof{figure}{
    \textbf{\ourmodel grounds heterogeneous modalities - vision, language, and action - in explicit 3D geometry.}  
    Multi-view visual features are spatially aligned and fused into point representations, while language conditioning highlights task-relevant semantics.
    For action generation, 3D spatial anchors connect action predictions with observed points in a shared coordinate frame, enabling spatially grounded action-to-point interactions.
    \ourmodel achieves state-of-the-art performance across diverse robotic manipulation benchmarks, despite having over 10$\times$ fewer trainable parameters than larger competing models.
    }
    \label{fig:teaser}
    \vspace{-0.5em}
\end{strip}

\begin{abstract}

While recent vision-language-action (VLA) models increasingly incorporate 3D geometry, how to effectively exploit it for precise and generalizable action prediction remains a central challenge. 
We introduce \textbf{\ourmodel}, a 3D policy that grounds \textbf{h}eterogeneous modal\textbf{i}ties - vision, language, and action - in explicit \textbf{3}D geomet\textbf{r}y for robotic manipulation. 
Our model integrates spatially-aligned visual features and language conditioning into point cloud representations, enriching scene geometry with fine-grained, task-relevant semantics.
It then anchors low-level control commands in the same 3D coordinate frame and uses 3D-RoPE enhanced action-to-point attention over coarse-to-fine point features to model geometric relationships between action targets and observed points.
Our analysis demonstrates the importance of semantic fusion within 3D and spatial alignment between observations and actions.
With fewer than 250M trainable parameters, \ourmodel achieves state-of-the-art performance on RoboCasa, RLBench, RoboTwin2, DexJoCo and DexSuite across diverse manipulation tasks and embodiments, from single-arm and dual-arm grippers to dexterous hands.
Real-robot evaluations further demonstrate its effectiveness in the physical world.
The model, code and dataset are publicly released.

\end{abstract}
    
\section{Introduction}
\label{sec:intro}

Language-guided robotic manipulation~\cite{guhur2023hiveformer,intelligence2025pi05,bjorck2025gr00t} requires both understanding task-relevant semantics and reasoning about physical geometry to generate precise actions.
Recent vision-language-action (VLA)~\cite{intelligence2025pi05,bjorck2025gr00t} models have demonstrated promising capabilities across diverse manipulation tasks, but their predominantly 2D visual representations do not well capture metric 3D structure. 
This has motivated the development of 3D VLAs~\cite{qu2025spatialvla,chen2026pointact,sun2025geovla} that incorporate explicit geometric observations, such as point clouds, to support robot action generation.

Existing 3D policies differ in how they represent geometry and combine it with vision and language for action prediction.
Some only associate pretrained 2D visual features with 3D coordinates~\cite{gervet2023act3d,ke20243dda,xiong2026geomvla},  
while others explicitly encode point clouds with dedicated 3D backbones~\cite{qi2017pointnet,wu2024ptv3,zhang2026utonia} or fuse the resulting geometric features with visual representations~\cite{fan2026any3dvla,chen2026pointact}.
The resulting features can be integrated into a shared VLM backbone~\cite{yang2026fp3,fan2026any3dvla} or fed into a separate action expert~\cite{sun2025geovla}.
PointACT~\cite{chen2026pointact} compares these architectures and highlights the benefits of an action expert with multi-scale point features.
Despite these advances, it remains unclear which design choices are most critical for mapping explicit geometry into effective action learning, and how well they apply across manipulation tasks and embodiments, from single-arm grippers to bimanual dexterous hands.
These questions motivate our investigation:
\emph{What makes explicit 3D geometry effective for action prediction, and how robust are these design choices across different manipulation settings?}

We introduce \ourmodel, a 3D-based policy that grounds heterogeneous modalities---vision, language, and action---in explicit 3D geometry, as illustrated in~\Cref{fig:teaser}. 
Our key idea is to use 3D point representations as a common geometric reference for multimodal perception and action generation. 
On the perception side, we integrate spatially-aligned image features and language conditioning into point representations, enriching scene geometry with fine-grained, task-relevant semantics. 
On the action side, a flow-matching action expert attends to these multimodal point representations in a coarse-to-fine manner. 
However, action representations, such as joint configurations or relative motion commands, may not be directly aligned with the 3D geometry of the observed scene.
To bridge this gap, we introduce spatial anchors that augment the original action representations with 3D end-effector positions.
These anchors lie in the scene coordinate frame, allowing 3D rotary positional encoding (RoPE)~\cite{su2024rope} in action-to-point attention to capture relative spatial relationships between action targets and observed points.
Additionally, we investigate the benefits of pretraining \ourmodel on a large-scale real-robot dataset DROID~\cite{khazatsky2024droid}. 

We extensively evaluate \ourmodel on multiple manipulation benchmarks, including RoboCasa24~\cite{nasiriany2024robocasa}, RLBench~\cite{james2020rlbench}, RoboTwin2~\cite{chen2025robotwin2},  DexJoCo~\cite{wang2026dexjoco}, and DexSuite~\cite{houssaini2026dexsuite}, which cover 
diverse
tasks and embodiments, from single-arm and dual-arm manipulators to dexterous hands.
\ourmodel achieves state-of-the-art performance across  benchmarks with over 10$\times$ fewer trainable parameters than state-of-the-art methods such as $\pi_{0.5}$~\cite{intelligence2025pi05}, even without pretraining on large-scale real-robot data.
Extensive ablations show that spatial anchors and 3D positional encoding are critical for precise action prediction by aligning actions with observed geometry, while fusing visual and language features into the point backbone provides gains beyond multimodal conditioning in the action expert.

Our contributions are summarized as follows:
\begin{itemize}
    \item We introduce \ourmodel, a 3D-based policy that uses explicit geometric representations to connect multimodal perception with spatially grounded action generation.
    
    \item We investigate key design choices in \ourmodel, including multimodal fusion, spatially grounded action prediction, and pretraining, providing empirical insights for building effective 3D policies.
        
    \item \ourmodel achieves state-of-the-art performance across diverse simulation benchmarks and real-world manipulation tasks with fewer than 250M trainable parameters.
\end{itemize}

\begin{figure*}[t]
    \centering
    \includegraphics[width=1\linewidth]{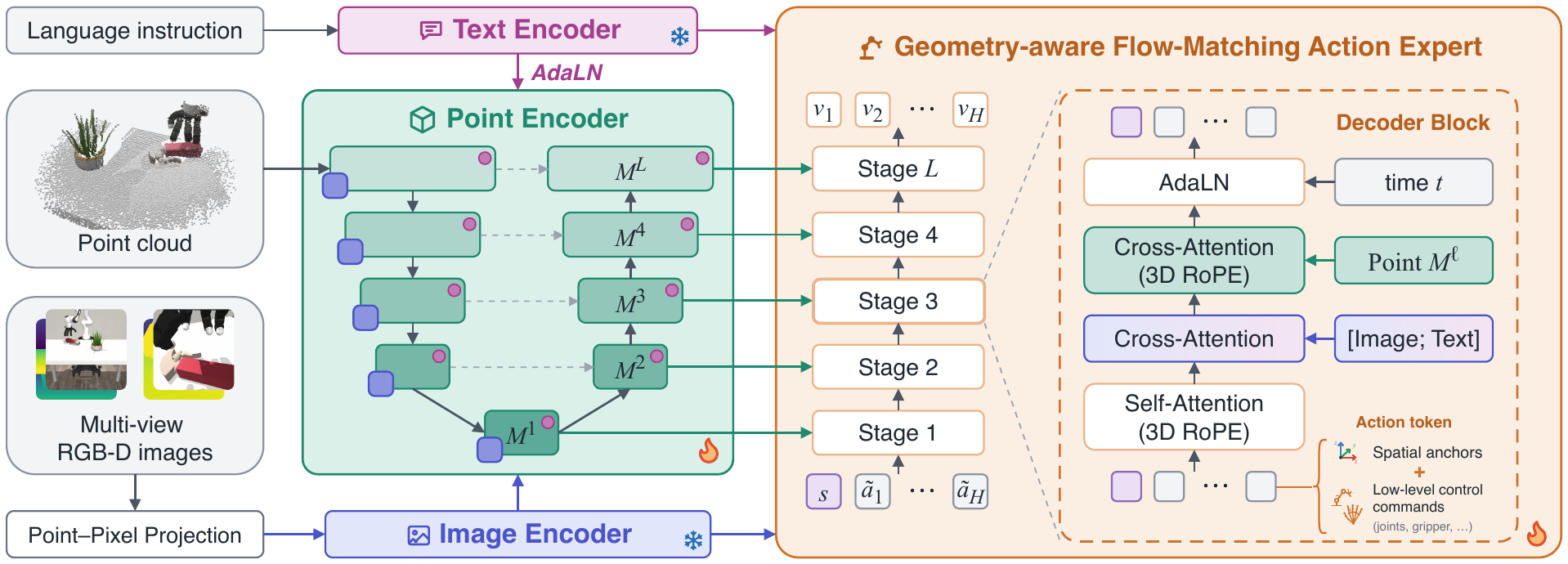}
    \caption{\textbf{\ourmodel architecture.} Multi-view image features are spatially aligned and fused with hierarchical point representations, while language features provide semantic conditioning through adaptive layer normalization.
    A flow-matching action expert attends to image and text tokens, as well as multi-scale point features in a coarse-to-fine manner, to predict action chunks.
    We augment the original action representations with spatial anchors encoding 3D end-effector positions, enabling 3D RoPE in both action self-attention and action-to-point attention to capture relative spatial relationships. 
    }
    \label{fig:method_arch}
    \vspace{-1em}
\end{figure*}

\section{Related Work}
\label{sec:related_work}

\noindent\textbf{Vision-language robotic manipulation.}
Vision-language-action (VLA) models such as RT-2~\cite{brohan2023rt2}, OpenVLA~\cite{kim2024openvla}, $\pi$~\cite{intelligence2025pi05,intelligence2026pi07} and GR00T~\cite{bjorck2025gr00t}, have advanced generalist manipulation by fine-tuning pretrained vision-language models (VLMs)~\cite{bai2025qwen3vl,beyer2024paligemma} on large-scale robot demonstrations.
More recent world-action models (WAMs)~\cite{ye2026dreamzero,kim2026cosmospolicy} further show benefits from future state prediction and video pretraining for robot control.
To better understand the design choices underlying these models, several works~\cite{kim2025openvlaoft,wu2026vlanext,wang2026openwam} examine visual representations, architectures, fine-tuning strategies, and training data.
While these studies provide valuable guidance for image- and video-based policies, our work examines how explicit 3D geometry can support policy learning, a question that remains comparatively underexplored.

\noindent\textbf{Geometric priors for visual policies.}
A growing body of work incorporates geometric knowledge into visual policies to improve spatial understanding and manipulation.
SpatialVLA~\cite{qu2025spatialvla} and SPEAR-1~\cite{nikolov2026spear1} augment visual representations with pretrained depth estimators, while E0.5~\cite{team2026e05} adopts geometric foundation models like VGGT~\cite{wang2025vggt} as policy backbones.
Other approaches solely introduce geometric supervision during training. %
SpatialForcing~\cite{li2026spatialforcing} aligns intermediate visual features with VGGT representations; 
Pri4R~\cite{kim2026pri4r} uses dense 3D point tracking as the auxiliary loss;
and SA-WAM~\cite{gonzalez2026sawam} incorporates geometry into WAMs through joint RGB, depth, and action prediction.
To better connect the observation and action space, some works first predict 3D intermediate representations such as 3D object pose~\cite{lin2026posevla}, camera-frame SE(3) end-effector trajectory~\cite{lin2026posevla,chen2026oasis} or 3D scene flows~\cite{wang2026lamp}, then condition an action expert on the resulting latent features.
However, 
this formulation does not align multi-view visual observations, and requires large-scale 3D pretraining~\cite{lin2026posevla,wang2026lamp}.
Our approach directly encodes metric point clouds aggregated from multiple views and uses spatial anchors to align action tokens with scene points in a common coordinate frame, enabling data- and compute-efficient action prediction.

\noindent\textbf{3D-based robotic manipulation.}
The prevalence of 3D sensors has encouraged explicit 3D policy learning.
Existing 3D-based policies differ in how they represent scene geometry and incorporate it into action prediction.
Rendering-based methods~\cite{goyal2023rvt,goyal2024rvt2,li2026bridgevla} project point clouds into virtual views,
allowing to reuse pretrained image models.
Depth- or point-map-based methods~\cite{jia2026pointmappolicy,lee2026seervla} represent geometry on RGB-aligned 2D grids, facilitating image-based encoding and fusion.
Another family constructs 3D feature clouds by lifting image features to 3D with their corresponding 3D coordinates~\cite{gervet2023act3d,ke20243dda,gkanatsios20253dflowmatchactor,xiong2026geomvla} and enhances action prediction via 3D rotary position embeddings~\cite{su2024rope}.
Due to improvements in 3D point encoders~\cite{qi2017pointnet,wu2024ptv3,zhang2026utonia}, more recent methods directly encode point clouds~\cite{chen2023polarnet,ze2024dp3,wang2024rise,garcia2025gembench} and inject them into VLAs.
Some feed point features into the VLM backbone~\cite{yang2026fp3,fan2026any3dvla}, whereas others~\cite{li2026pointvla,sun2025geovla,li2026bridge3d} incorporate them into the action expert.
PointACT~\cite{chen2026pointact} compares these two architectures and shows the benefits of a separate action expert conditioned on multi-scale point features.
PointWAM~\cite{park2026pointwam} extends point-based modeling to cascaded world action models for dexterous manipulation.
In this work, we introduce \ourmodel, which improves spatially grounded multimodal perception and action generation, and demonstrate its effectiveness across diverse manipulation benchmarks and robot embodiments.

\section{Method}
\label{sec:method}

\subsection{\ourmodel Architecture}

We consider language-guided robotic manipulation, where a policy predicts a sequence of robot actions from visual observations, 3D scene geometry, and a natural language instruction. 
At each timestep $t$, the robot receives a point cloud $\mathcal{P}_t$, RGB images $\mathcal{I}_t = \{I_t^{v}\}_{v=1}^{V}$ from $V$ camera views, a proprioceptive state $\mathbf{s}_t$, and a language instruction $\mathcal{L}$.
We construct $\mathcal{P}_t$ from calibrated RGB-D cameras, establishing spatial correspondences between scene points and image pixels.
The goal is to predict an action chunk $\mathbf{A}_t = (\mathbf{a}_t,\ldots,\mathbf{a}_{t+H-1})$ of horizon $H$.

As illustrated in~\Cref{fig:method_arch}, \ourmodel consists of multimodal encoders and a geometry-aware action expert.

\noindent\textbf{Image and text encoders.}
We use frozen SigLIP2~\cite{tschannen2025siglip2} encoders to extract visual and language features.
Each camera view is encoded independently into a spatial feature map $\mathbf{V}^{v}_t$.
We flatten and concatenate these maps across views to form a visual token sequence $\mathbf{V}_t \in \mathbb{R}^{N_V \times d_V}$.
The text encoder produces language embeddings $\mathbf{T} \in \mathbb{R}^{N_T \times d_T}$.
These features support both multimodal fusion within the point encoder and direct conditioning of the action expert.

\noindent\textbf{Multimodal point encoder.}
We use a multimodal point encoder $\mathcal{E}_{P}$ built on Point Transformer v3 (PTv3)~\cite{wu2024ptv3,zhang2026utonia}.
Given the point cloud $\mathcal{P}_t$, it integrates spatially aligned visual features $\mathbf{V}_t$ and language conditioning $\mathbf{T}$ to produce task-conditioned point representations at multiple scales:
\begin{equation}
\{\mathcal{M}^{\ell}\}_{\ell=1}^{L}
=
\mathcal{E}_{P}(\mathcal{P}_t, \mathbf{V}_t, \mathbf{T}).
\end{equation}
The backbone follows a hierarchical encoder-decoder architecture: the encoder progressively aggregates local geometric information, while the decoder recovers finer spatial detail through upsampling and skip connections.
We collect representations from $L$ decoder stages, where
\begin{equation}
\mathcal{M}^{\ell}
=
\{(\mathbf{p}^{\ell}_i, \mathbf{f}^{\ell}_i)\}_{i=1}^{N_\ell},
\qquad
\mathbf{p}^{\ell}_i \in \mathbb{R}^{3},
\quad
\mathbf{f}^{\ell}_i \in \mathbb{R}^{d_\ell}.
\end{equation}
Here, $\mathbf{p}^{\ell}_i$ and $\mathbf{f}^{\ell}_i$ denote the 3D coordinates and multimodal features of point $i$ at scale $\ell$.
The representations range from the coarsest scale $\mathcal{M}^{1}$ to the finest scale $\mathcal{M}^{L}$.

\noindent\textbf{Geometry-aware action expert.}
We use a conditional flow-matching action expert that transforms sampled noise into an action chunk.
It attends to the multi-scale point representations in a coarse-to-fine manner, with image and language features providing complementary conditioning.
Spatial anchors connect action tokens to scene geometry in a shared coordinate frame, enabling spatially grounded action-to-point attention.

We next detail the multimodal point encoder and the geometry-aware action expert.

\subsection{Grounding Multimodal Semantics in 3D}

Point clouds provide explicit metric geometry, but their sparsity limits the fine-grained appearance.
We therefore fuse spatially-aligned image features into point representations to incorporate richer visual information from pretrained image encoders.
We further introduce language conditioning to emphasize geometry relevant to the task goal.

\noindent\textbf{Point-image fusion.}
We associate image features with scene points using calibrated camera parameters.
For each point $\mathbf{p}_i$ and camera view $v$, we extract its visual feature as
$\mathbf{g}_i^{v} = \operatorname{Bilinear}(\mathbf{V}^{v}, \operatorname{proj}_v(\mathbf{p}_i))$,
where $\operatorname{proj}_v$ maps 3D coordinates to image coordinates and $\operatorname{Bilinear}$ samples the corresponding feature map.
We retain projections that satisfy image-boundary and depth-consistency constraints, average their features across valid views as the per point image feature $\mathbf{g}_i$.
Unlike PointWorld~\cite{huang2026pointworld} which only concatenates $\mathbf{g}_i$ with point features at the encoder input, we inject aligned visual features at every encoder stage.
We propagate these features to coarser scales by average pooling over the corresponding point downsampling groups and fuse them with the point features at each stage.
This enables visual semantic information to be incorporated into geometric representations at different spatial resolutions.

\noindent\textbf{Point-text fusion.}
We condition the point cloud backbone on language to emphasize task-relevant scene geometry.
Learnable queries are first used to aggregate the language token embeddings into a global representation $\mathbf{c}_T$ through attentive pooling.
We then modulate point features using adaptive layer normalization (AdaLN)~\cite{peebles2023dit}:
\begin{equation}
    \operatorname{AdaLN}(\mathbf{f}^{\ell}_i;\mathbf{c}_T)
    =
    (1+\boldsymbol{\gamma}^{\ell})
    \odot \operatorname{LN}(\mathbf{f}^{\ell}_i)
    + \boldsymbol{\beta}^{\ell},
    \label{eq:text_adaln}
\end{equation}
where $(\boldsymbol{\gamma}^{\ell},\boldsymbol{\beta}^{\ell})
= \operatorname{MLP}_{\ell}(\mathbf{c}_T)$ are layer-specific scale and shift parameters shared across points.
This enables efficient task-dependent modulation of point features.

The resulting hierarchical features
$\{\mathcal{M}^{\ell}\}_{\ell=1}^{L}$
encode geometry, visual appearance, and task-relevant semantics at multiple spatial scales for subsequent action generation.

\subsection{Grounding Action Generation in 3D}

We employ a transformer-based flow-matching action expert, predicting the velocity field that transforms sampled noise into an action chunk given multimodal inputs.

\noindent\textbf{Coarse-to-fine multimodal conditioning.}
Each decoder block in the action expert combines action-token self-attention with cross-attention to point, image, and language features.
To exploit hierarchical point representations efficiently, we organize the decoder blocks into groups that attend to progressively finer point features, from $\mathcal{M}^{1}$ to $\mathcal{M}^{L}$.
This coarse-to-fine schedule first captures broad scene context and then incorporates local geometric detail.
Direct conditioning on image and language embeddings provides complementary semantic information.

\begin{figure*}[t]
  \centering
  \begin{subfigure}[t]{0.3263\textwidth}
    \captionsetup{margin={0.1635\linewidth,0pt}}
    \includegraphics[width=\linewidth]{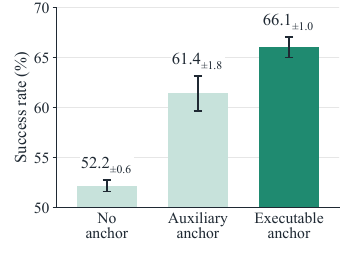}
    \caption{Spatial anchor.}
    \label{fig:ablation_spatial_anchor}
  \end{subfigure}%
  \hfill
  \begin{subfigure}[t]{0.3688\textwidth}
    \includegraphics[width=\linewidth]{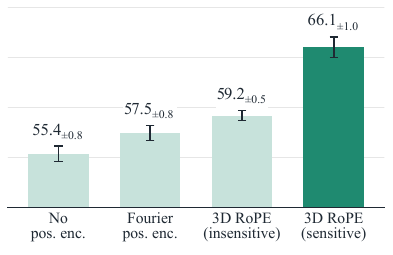}
    \caption{3D position encoding.}
    \label{fig:ablation_3dposenc}
  \end{subfigure}%
  \hfill
  \begin{subfigure}[t]{0.2749\textwidth}
    \includegraphics[width=\linewidth]{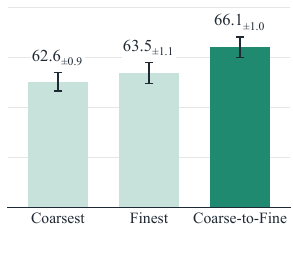}
    \caption{Point features.}
    \label{fig:ablation_point_features}
  \end{subfigure}
  \caption{Ablation of the geometry-aware action expert on RoboCasa24. Results are reported as mean success rates over three seeds.}
  \label{fig:action_expert_ablation}
  \vspace{-1em}
\end{figure*}

\noindent\textbf{Spatial anchors for 3D positional encoding.}
Precise manipulation requires reasoning about the relative geometry between action targets and observed scene points.
However, robot actions may be expressed as joint configurations or relative motion commands, which do not directly specify positions in the scene's 3D coordinate frame.
We therefore augment action representations with \emph{spatial anchors}, typically the corresponding end-effector positions expressed in the same coordinate frame as the point cloud.
These anchors provide a geometric reference for attention while retaining robot-specific action parameterization.

We use spatial anchors to apply 3D rotary positional encoding (3D RoPE)~\cite{su2024rope,gervet2023act3d} to action self-attention and action-to-point cross-attention.
The anchors and points are expressed in metres in a shared coordinate frame.
The rotary transformation $R(\cdot)$ partitions feature dimensions among the three spatial axes and rotates each feature pair according to its assigned frequency and the coordinate along that axis.
Specifically, for coordinate $u$, the $m$-th feature pair is transformed by
\begin{equation}
    R_m(u) =
    \begin{bmatrix}
        \cos(\omega_m u) & -\sin(\omega_m u) \\
        \sin(\omega_m u) &  \cos(\omega_m u)
    \end{bmatrix},
    \omega_m = \omega_{\max}\theta^{-2m/d_r},
    \label{eq:rope_rotation}
\end{equation}
where $d_r$ is the rotary dimension per axis and
$m \in \{0,\ldots,d_r/2-1\}$.
The full transformation combines these rotations across all feature pairs and spatial axes.

For an action query $\mathbf{q}_i$ with current noisy anchor $\mathbf{x}_i$ and a point key $\mathbf{k}_j$ at $\mathbf{p}_j$, the resulting attention score is
\begin{equation}
    s_{ij}
    =
    \frac{
        \bigl(R(\mathbf{x}_i)\mathbf{q}_i\bigr)^\top
        \bigl(R(\mathbf{p}_j)\mathbf{k}_j\bigr)
    }{\sqrt{d}}
    =
    \frac{
        \mathbf{q}_i^\top
        R(\mathbf{p}_j-\mathbf{x}_i)\mathbf{k}_j
    }{\sqrt{d}},
    \label{eq:3d_rope}
\end{equation}
where $d$ is the attention head dimension.
This makes attention explicitly sensitive to relative spatial displacement.
Action self-attention follows the same formulation, replacing $\mathbf{p}_j$ with the spatial anchor of the key action token.
The frequency spectrum controls the spatial sensitivity of the encoding: a displacement $\Delta u$ induces a rotation of $\omega_m\Delta u$.
Higher frequencies provide greater sensitivity to small displacements, while lower frequencies vary more slowly over larger distances.
The maximum frequency $\omega_{\max}$ sets the finest spatial scale, while the base $\theta$ controls how far the frequencies spread toward coarser scales, allowing the encoding to capture relative geometry at multiple spatial scales.

\subsection{Training and Inference}

\paragraph{Flow-matching action learning.}
We train the action expert using conditional flow matching.
Given a ground-truth action chunk $\mathbf{A}_t$, we sample
Gaussian noise $\boldsymbol{\epsilon}\sim\mathcal{N}(0,\mathbf{I})$
and a flow timestep $\tau \in [0, 1]$,
and construct the interpolated actions
$\mathbf{A}_t^\tau=(1-\tau)\boldsymbol{\epsilon}+\tau\mathbf{A}_t$.
Conditioned on the observations
$\mathcal{O}_t=(\mathcal{P}_t,\mathcal{I}_t,\mathbf{s}_t,\mathcal{L})$,
including the point cloud, images, robot state, and language instruction,
the model predicts a velocity field $v_\theta$ by minimizing
\begin{equation}
    \mathcal{L}_{\mathrm{FM}} =
    \mathbb{E}_{\tau,\boldsymbol{\epsilon}}
    \left[
    \left\|
    v_\theta(\mathbf{A}_t^\tau,\tau,\mathcal{O}_t)
    -(\mathbf{A}_t-\boldsymbol{\epsilon})
    \right\|_2^2
    \right].
    \label{eq:flow_matching}
\end{equation}
During inference, we generate an action chunk by solving
the ODE
$d\mathbf{A}_t^\tau/d\tau =
v_\theta(\mathbf{A}_t^\tau,\tau,\mathcal{O}_t)$,
starting from Gaussian noise at $\tau=0$ and integrating to $\tau=1$.

\noindent\textbf{Implementation Details.}
We use $L=5$ stages in the point encoder.
The action expert assigns 3 decoder blocks per stage, containing 15 blocks in total, each with a hidden size of 512 and 8 attention heads.
For 3D RoPE, we set $\omega_{\max}=50$\,rad/m and $\theta=100$, which with $d_r=20$ gives frequencies from 50 to 0.79\,rad/m, i.e., wavelengths from 12.6\,cm to 7.9\,m.
We initialize the point encoder from the self-supervised Utonia PTv3 backbone~\cite{zhang2026utonia} and fine-tune it during training.
The SigLIP2~\cite{tschannen2025siglip2} image and text encoders remain frozen except for their final projection layers, resulting in fewer than 250M trainable parameters overall.
We train with the AdamW optimizer, using a learning rate of $10^{-4}$ and a cosine schedule.
Training uses a global batch size of 128 on 2 NVIDIA H100 GPUs. %
At inference, actions are generated with 10 Euler integration steps.
\Cref{app:implementation} of the supplementary material provides more details.

\noindent\textbf{Pretraining on DROID.}
To improve generalization, we pretrain \ourmodel on DROID~\cite{khazatsky2024droid}, a real-world dataset of teleoperated robot demonstrations of a single-arm gripper.
Since the original DROID dataset lacks depth images and contains noise in camera calibration and language annotations, we construct an RGB-D version using 3D reconstructions from PointWorld~\cite{huang2026pointworld} and curated language annotations from RoboInter~\cite{li2026robointer}, including 35.2k demonstrations with 25.4k distinct instructions.
Each frame has two calibrated external RGB-D views and a wrist-camera RGB image.
Further details are presented in \Cref{app:droid} of the supplementary material.

\section{Ablation Studies}
\label{sec:ablation_expr}

This section performs controlled ablations of \ourmodel to identify key design choices for effective 3D policy learning.

\noindent\textbf{Experimental setup.}
We use RoboCasa24~\cite{nasiriany2024robocasa} as our primary ablation benchmark, which comprises 24 tasks spanning diverse manipulation skills and testing generalization across environments and objects.
We train a multi-task policy for $\sim$60k iterations, using 50 human teleoperation demonstrations per task.
Following prior work~\cite{kim2026cosmospolicy}, we evaluate each task over 50 episodes per seed across three random seeds and report the mean success rate and standard deviation across seeds.
All ablations in this section are conducted without DROID pretraining. 
Additional ablations on the dexterous manipulation benchmark DexJoCo~\cite{wang2026dexjoco} and the benefits of DROID pretraining are presented in \Cref{app:dexjoco_ablation} and \ref{app:pretrain_ablation}, respectively, in the supplementary material.

\subsection{How to Ground Actions in 3D Geometry?}

To isolate the contribution of 3D geometry, we remove the image encoder and condition the action expert solely on point cloud and language features in this subsection.

\noindent\textbf{Spatial anchors.}
We compare three variants to investigate how spatial anchors affect action prediction.
\begin{itemize}
    \item \emph{Executable anchor}: Absolute end-effector positions are predicted, followed by the remaining control command. The absolute action is used in 3D RoPE and action execution.
    Since RoboCasa24 uses delta control, predicted absolute positions from executable anchors are converted into positional deltas and combined with other control commands like gripper openness for execution.
    \item \emph{Auxiliary anchor}: The control action is augmented with a prefix whose training target is the next end-effector position. The anchor is only used for 3D RoPE, whereas the robot executes the predicted low-level commands.
    \item \emph{No anchor}: Only native control commands are predicted, and therefore action-side 3D RoPE is omitted.
\end{itemize}
As shown in~\Cref{fig:ablation_spatial_anchor}, the auxiliary anchor improves the success rate by 9.2 points over the no-anchor variant, demonstrating the benefits of explicitly grounding actions in the 3D scene, even when the spatial anchors are not directly executed.
Executable anchors further improve performance, suggesting that directly predicting action targets in the same coordinate frame as scene geometry enables more effective spatially grounded action generation.

\noindent\textbf{3D positional encoding.}
We investigate how different 3D positional encoding strategies affect spatial interactions between action queries and scene points:
\begin{itemize}
    \item \emph{3D RoPE}, which encodes relative spatial relationships in \cref{eq:3d_rope}. We examine the effect of its frequency parameters in \cref{eq:rope_rotation}, which control the sensitivity of the encoding to spatial displacements. The insensitive version uses $\omega_{\text{max}}=1, \theta=10$ following Utonia~\cite{zhang2026utonia}.
    \item \emph{Fourier features}, which map coordinates to multi-frequency sinusoidal features, followed by an MLP that projects them into positional embeddings added to the corresponding features.
    \item \emph{No positional encoding}, which does not explicitly incorporate 3D coordinates into the action expert's attention.
\end{itemize}
\Cref{fig:ablation_3dposenc} shows that 3D RoPE outperforms Fourier features and the no-encoding baseline, highlighting the importance of explicitly modeling relative spatial relationships between actions and scene geometry.
Furthermore, 3D RoPE (sensitive), which uses higher frequencies to better distinguish nearby positions, achieves significantly better performance than the insensitive variant, suggesting that fine-grained spatial discrimination is crucial for precise action generation.

\noindent\textbf{Multi-scale point features.}
We compare conditioning the action expert on a single feature scale against a coarse-to-fine strategy that progressively incorporates finer-scale features across decoder blocks.
As shown in~\Cref{fig:ablation_point_features}, fine-grained features slightly outperform coarse features, while coarse-to-fine conditioning achieves the best performance.
This suggests that combining global scene context with fine-grained geometric details benefits action prediction while keeping attention costs manageable.

\subsection{How to Integrate Vision, Language and 3D?}

\begin{table}[t]
  \centering
  \caption{
  Ablation of multimodal fusion in point encoder and conditioning in action expert on RoboCasa24. 
  }
  \label{tab:multimodal_fusion_ablation}
  \vspace{-0.5em}
  \begin{tabular}{cccc}
    \toprule
    Point-text & Point-image & Action-image & \multirow{2}{*}{SR (\%)} \\
    fusion     & fusion      & condition    & \\
    \midrule
    $\times$   & $\times$    & $\times$     & 57.6$_{\pm 0.8}$ \\
    Cross-attn & $\times$    & $\times$     & 63.1$_{\pm 1.4}$ \\
    AdaLN      & $\times$    & $\times$     & 66.1$_{\pm 1.0}$ \\
    \midrule
    \multirow{4}{*}{AdaLN} & Input       & $\times$     & $69.3_{\pm 0.6}$ \\
               & All stages  & $\times$     & 71.9$_{\pm 0.9}$ \\
               & $\times$ & $\checkmark$ & 67.9$_{\pm 0.2}$ \\
               & All stages  & $\checkmark$ & \textbf{72.3}$_{\pm 0.6}$ \\
    \bottomrule
  \end{tabular}
  \vspace{-1em}
\end{table}

We explore where and how to integrate heterogeneous modalities into the point encoder and action expert.

\noindent\textbf{Point-text fusion.}
In ~\Cref{tab:multimodal_fusion_ablation} (upper block), we investigate whether incorporating language into point representations provides additional benefits beyond direct language conditioning in the action expert.
We compare three variants: no point-text fusion, fusion with a pooled text embedding via AdaLN, and token-level fusion through cross-attention.
Incorporating language into point representations consistently improves policy performance, even when the action expert already has access to language, demonstrating the necessity of learning task-relevant 3D representations.
In addition, AdaLN outperforms token-level cross-attention with lower computational overhead.
We hypothesize that the instructions primarily specify task goals and provide limited fine-grained object descriptions, so a global instruction embedding may capture the task semantics needed to modulate 3D point features.
A related observation is reported in DiT~\cite{peebles2023dit}, where AdaLN outperforms cross-attention for class-conditional image generation.

\begin{table}[t]
    \centering
    \caption{
    Performance on the RoboCasa24 benchmark.
    Model size is the number of trainable parameters. 
    }
    \label{tab:robocasa24_sota}
    \vspace{-0.5em}
    \resizebox{\columnwidth}{!}{%
    \begin{tabular}{@{}lccc@{}}
        \toprule
        Model
        & \shortstack{Model\\size}
        & \shortstack{Training demos\\per task}
        & \shortstack{Average\\SR (\%)} \\
        \midrule
        \rowcolor{gray!12} \multicolumn{4}{@{}l}{\textit{2D-based policies}} \\
        \quad GR00T-N1~\cite{bjorck2025gr00t}                 & 2.2B     & 300 & 49.6 \\
        \quad $\pi_0$~\cite{black2024pi0}                     & 3.1B     & 300 & 62.5 \\
        \quad GR00T-N1.5~\cite{bjorck2025gr00t}               & 2.7B     & 300 & 64.1 \\
        \quad Video Policy~\cite{liang2025videopolicy}        & $>$1.5B       & 300 & 66.0 \\
        \quad GR00T-N1.5 + HAMLET~\cite{koo2026hamlet}        & $>$2.7B  & 300 & 66.4 \\
        \quad $\pi_{0.5}$~\cite{intelligence2025pi05}         & 3.2B     & 50  & 52.9 \\
        \quad Cosmos Policy~\cite{kim2026cosmospolicy}        & 2B       & 50  & 67.1 \\
        \midrule
        \rowcolor{gray!12} \multicolumn{4}{@{}l}{\textit{3D-aware policies}} \\
        \quad FP3~\cite{yang2026fp3}                          & 1.3B     & 50  & 42.9$_{\pm 0.7}$ \\
        \quad $\pi_{0.5}$ + Pri4R~\cite{kim2026pri4r}         & $>$3.2B  & 50  & 57.0 \\
        \quad $\pi_{0.5}$ + GeoVLA~\cite{sun2025geovla}       & $>$3.2B  & 50  & 57.4$_{\pm 1.4}$ \\
        \quad $\pi_{0.5}$ + PointVLA~\cite{li2026pointvla}    & $>$3.2B  & 50  & 58.0$_{\pm 0.7}$ \\
        \quad $\pi_{0.5}$ + SeeR-VLA~\cite{lee2026seervla}    & $>$3.2B  & 50  & 61.5$_{\pm 1.2}$ \\
        \quad SA-WAM~\cite{gonzalez2026sawam}       & 2B       & 50  & 76.6$_{\pm 0.5}$ \\
        \rowcolor{lightgreen}
        \quad \ourmodel (Ours)                                      & 0.25B       & 50  & $\textbf{78.8}_{\pm 0.8}$ \\ 
        \bottomrule
    \end{tabular}%
    }
    \vspace{-1.em}
\end{table}

\noindent\textbf{Image-point fusion.}
\Cref{tab:multimodal_fusion_ablation} also examines whether and where to fuse spatially aligned image features into the point encoder.
Image-point fusion substantially improves performance, with multi-stage fusion (+5.8) outperforming input-only fusion (+3.2) compared to the non-fusion baseline.
This demonstrates the benefits of integrating visual semantics with explicit 3D geometry before action generation.

\noindent\textbf{Action-level image conditioning.}
We further evaluate direct image conditioning in the action expert in~\Cref{tab:multimodal_fusion_ablation}.
It improves the success rate by 1.8 points without image-point fusion, but provides only a marginal gain of 0.4 points with multi-stage fusion.
This suggests that spatially-aligned image-point fusion is more effective than action-level image conditioning alone.
Nevertheless, direct image conditioning does not require explicit spatial alignment obtained from RGB-D cameras, making it valuable when geometric information is unavailable or inaccurate.
For instance, the wrist camera in DROID~\cite{khazatsky2024droid} provide useful visual information but lack reliable depth for image-point fusion. We therefore retain direct image conditioning in our final model.

\section{Comparison with State of the Art}
\label{sec:expr_sota}

We evaluate \ourmodel on five robotic manipulation benchmarks covering diverse tasks and robot embodiments, ranging from single-arm to bimanual coordination and high-DoF dexterous control, as well as real robots.
\ourmodel is first pretrained on DROID and then fine-tuned on each benchmark.
Benchmark descriptions and detailed results are provided in \Cref{app:benchmarks} and \ref{app:additional_results}, respectively, in the supplementary material.

\definecolor{bar2d}{HTML}{C9CDD3}
\definecolor{bar3d}{HTML}{A9D4C2}
\definecolor{barours}{HTML}{3B8A6E}
\definecolor{txtours}{HTML}{1F4D3D}
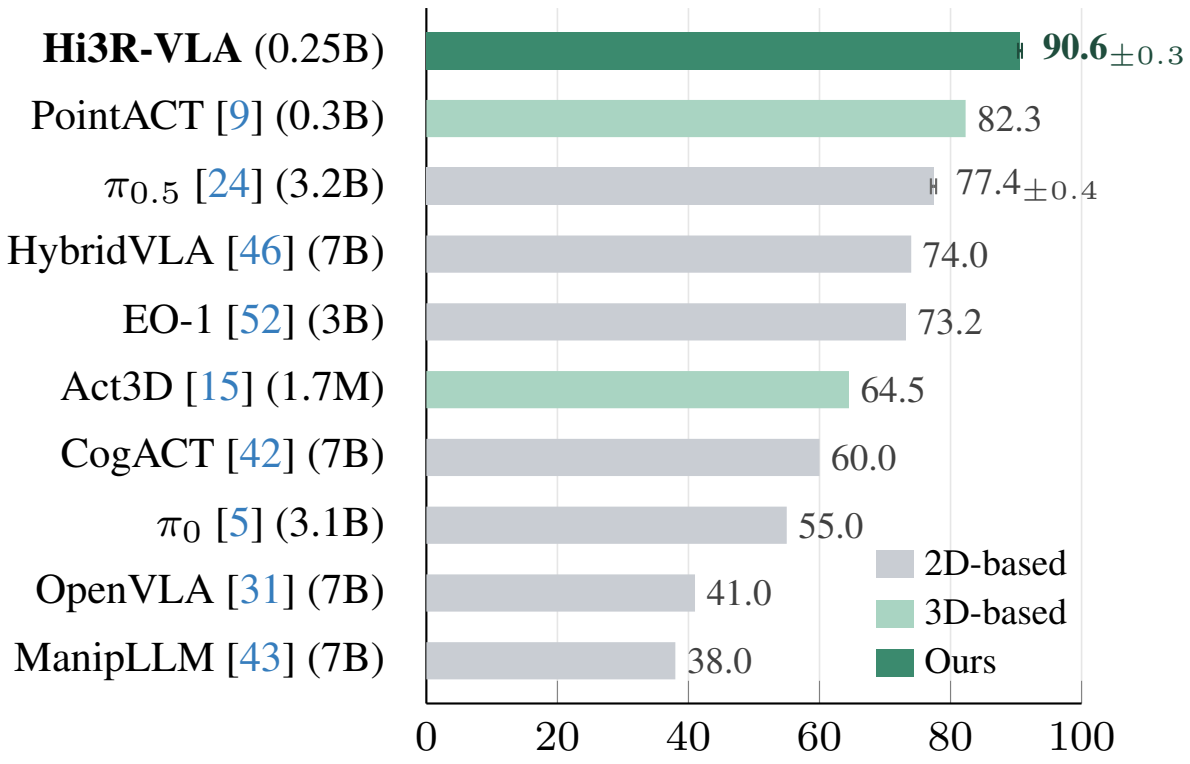
\begin{figure}[t]
    \centering
    \begin{tikzpicture}
    \begin{axis}[
        xbar, bar width=7pt, bar shift=0pt,
        width=0.52\columnwidth, height=4.6cm, scale only axis,
        xmin=0, xmax=100, xtick={0,20,...,100},
        xmajorgrids, major grid style={draw=black!10, line width=0.3pt},
        symbolic y coords={manip,openvla,pi0,cogact,act3d,eo1,hybrid,pi05,pointact,ours},
        ytick={manip,openvla,pi0,cogact,act3d,eo1,hybrid,pi05,pointact,ours}, enlarge y limits=0.07,
        yticklabels={
            ManipLLM~\cite{li2024manipllm} (7B),
            OpenVLA~\cite{kim2024openvla} (7B),
            $\pi_0$~\cite{black2024pi0} (3.1B),
            CogACT~\cite{li2024cogact} (7B),
            Act3D~\cite{gervet2023act3d} (1.7M),
            EO-1~\cite{qu2025eo1} (3B),
            HybridVLA~\cite{liu2025hybridvla} (7B),
            $\pi_{0.5}$~\cite{intelligence2025pi05} (3.2B),
            PointACT~\cite{chen2026pointact} (0.3B),
            \textbf{\ourmodel} (0.25B)},
        y tick label style={font=\footnotesize},
        x tick label style={font=\footnotesize},
        label style={font=\footnotesize},
        major y tick style={draw=none},
        axis x line*=bottom, axis y line*=left, axis line style={line width=0.4pt},
        legend style={at={(1.02,0.0)}, anchor=south east, draw=none, fill=none,
                      font=\scriptsize, row sep=-1pt},
        legend image code/.code={\fill[#1, draw opacity=0] (0cm,-0.08cm) rectangle (0.25cm,0.1cm);},
        legend cell align=left,
        clip=false,
    ]
    \addplot[fill=bar2d, draw=none,
             nodes near coords, nodes near coords style={font=\scriptsize, color=black!75, anchor=west, inner sep=2pt},
             point meta=explicit symbolic]
        coordinates {(38.0,manip)[38.0] (41.0,openvla)[41.0] (55.0,pi0)[55.0]
                     (60.0,cogact)[60.0] (73.2,eo1)[73.2] (74.0,hybrid)[74.0]};
    \addlegendentry{2D-based}
    \addplot[fill=bar2d, draw=none, forget plot,
             error bars/.cd, x dir=both, x explicit,
             error bar style={color=black!60, line width=0.6pt},
             error mark options={rotate=90, mark size=1.5pt, color=black!60}]
        coordinates {(77.4,pi05) +- (0.4,0)};
    \node[anchor=west, font=\scriptsize, text=black!75] at (axis cs:78.0,pi05)
        {77.4$_{\pm 0.4}$};
    \addplot[fill=bar3d, draw=none,
             nodes near coords, nodes near coords style={font=\scriptsize, color=black!75, anchor=west, inner sep=2pt},
             point meta=explicit symbolic]
        coordinates {(64.5,act3d)[64.5] (82.3,pointact)[82.3]};
    \addlegendentry{3D-based}
    \addplot[fill=barours, draw=none,
             error bars/.cd, x dir=both, x explicit,
             error bar style={color=txtours, line width=0.6pt},
             error mark options={rotate=90, mark size=1.5pt, color=txtours}]
        coordinates {(90.6,ours) +- (0.3,0)};
    \addlegendentry{Ours}
    \node[anchor=west, font=\scriptsize, text=txtours] at (axis cs:91.2,ours)
        {\textbf{90.6}$_{\pm 0.3}$};
    \end{axis}
    \end{tikzpicture}
    \vspace{-0.8em}
    \caption{
        Performance on the RLBench-10Task benchmark.
    }
    \label{fig:rlbench_10_sota}
    \vspace{-0.5em}
\end{figure}

\begin{table}[t]
\centering
\small
\setlength{\tabcolsep}{4pt}
\caption{Performance on the RoboTwin2 benchmark (50 tasks).
}
\label{tab:robotwin2_sota}
\vspace{-0.5em}
\begin{tabular}{lcccc}
\toprule
Method & Size & \makecell{clean$\rightarrow$ \\ clean} & \makecell{clean$\rightarrow$ \\ randomized} & Avg \\
\midrule
\rowcolor{gray!12} \multicolumn{5}{l}{\textit{2D-based policies}} \\
\quad Fast-WAM~\cite{yuan2026fastwam} & 6B & 77.8 & 1.9 & 39.8 \\
\quad Xiaomi-Robotics-0~\cite{cai2026xiaomi0} & 4.7B & 62.9 & 18.2 & 40.5 \\
\quad ABot-M0~\cite{yang2026abotm0} & 4.2B & 57.4 & 30.4 & 43.9 \\
\quad X-VLA~\cite{zheng2026xvla} & 0.9B & 68.0 & 20.9 & 44.5 \\
\quad Cosmos3-Nano~\cite{agarwal2026cosmos3} & 16B & 77.1 & 22.5 & 49.8 \\
\quad LingBot-VA~\cite{li2026lingbotva} & 5.3B & 80.7 & 34.6 & 57.7 \\
\quad $\pi_{0.5}$~\cite{intelligence2025pi05} & 3.2B & 70.7 & 46.0 & 58.3 \\
\midrule
\rowcolor{gray!12} \multicolumn{5}{l}{\textit{3D-aware policies}} \\
\quad X-WAM~\cite{guo2026xwam} & $>$5B & 70.0 & 25.8 & 47.9 \\
\quad SpatialForcing~\cite{li2026spatialforcing} & 3.2B & 77.2 & 26.7 & 52.0 \\
\quad 4D-WAM~\cite{yang20264dwam} & 5.3B & 81.5 & 41.8 & 61.6 \\
\rowcolor{lightgreen}
\quad \ourmodel (Ours) & 0.25B & \textbf{90.1} & \textbf{50.5} & \textbf{70.3} \\
\bottomrule
\end{tabular}
\vspace{-1.3em}
\end{table}

\noindent\textbf{RoboCasa24.}
Following the experimental setup in~\Cref{sec:ablation_expr}, we compare \ourmodel with state-of-the-art methods on RoboCasa24~\cite{nasiriany2024robocasa}.
As shown in~\Cref{tab:robocasa24_sota}, \ourmodel significantly outperforms CosmosPolicy~\cite{kim2026cosmospolicy}, the strongest 2D baseline by 11.7 points.
Our model also performs better than SA-WAM~\cite{gonzalez2026sawam}, a 3D-enhanced method built upon CosmosPolicy, using around $10\times$ fewer trainable parameters and substantially less pretraining data.

\noindent\textbf{RLBench.}
We follow the RLBench~\cite{james2020rlbench} 10-task evaluation protocol~\cite{chen2026pointact}, which comprises 10 tabletop single-arm manipulation tasks.
Unlike benchmarks that predict microstep control commands, RLBench uses keystep actions, requiring the policy to predict precise end-effector targets over larger spatial displacements.
The test scenes introduce novel object placements and distractors. %
We train a single multi-task policy on 100 demonstrations per task for 42k iterations, and evaluate 100 episodes per task across three random seeds.
As shown in~\Cref{fig:rlbench_10_sota}, \ourmodel achieves an average success rate over 90\%, outperforming the previous best method~\cite{chen2026pointact}, by 8 points.
These results highlight the advantages of \ourmodel for long-range keystep action prediction and spatial generalization.

\noindent\textbf{RoboTwin2.}
RoboTwin2~\cite{chen2025robotwin2} evaluates bimanual robotic manipulation across 50 tasks, including object rearrangement, tool use, and coordinated dual-arm actions.
Following the official benchmark protocol, we train a single multi-task policy on 50 clean demonstrations per task for 90k iterations and evaluate it in both clean and randomized environments with 100 episodes per task.
The latter introduces variations in background, lighting, table height, distractor objects, and unseen instructions.
To accommodate bimanual manipulation, we introduce arm-specific embeddings to distinguish left and right arms, doubling the number of action tokens while retaining the same action expert architecture.
As shown in~\Cref{tab:robotwin2_sota}, \ourmodel substantially outperforms prior methods in clean environments and also improves under the randomized setup, demonstrating the effectiveness of our model for coordinated bimanual manipulation.
We leave pretraining on larger and more diverse datasets to future work to further improve generalization to randomized environments.

\begin{figure*}
    \centering
    \includegraphics[width=1\linewidth]{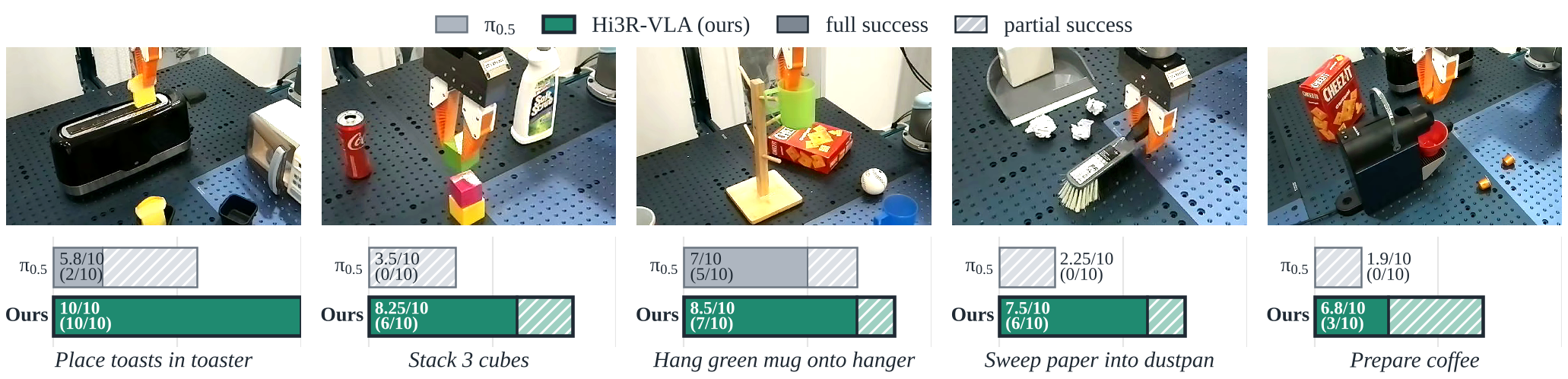}
    \vspace{-1.8em}
    \caption{\textbf{Real-robot results.} We compare $\pi_{0.5}$ and \ourmodel on five real-world tasks, with 10 testing trials per task. The full bar length is the partial-success score; the solid segment is the number of fully successful trials. \ourmodel outperforms $\pi_{0.5}$ on every task in both partial and full success, with particularly large gains on precise manipulation tasks such as \textit{Place toasts in toaster} and \textit{Prepare coffee}.
    }
\label{fig:real_robot_results}
\vspace{-1em}
\end{figure*}

\begin{table}[t]
    \centering
    \small
    \caption{
        Performance on DexJoCo (10 tasks) with Allegro hands.
    }
    \label{tab:dexjoco_sota_grouped}
    \vspace{-0.5em}
    \setlength{\tabcolsep}{4pt}
    \begin{tabular}{@{}lcccc@{}}
        \toprule
        Method & Size & Single-arm & Bimanual & All \\
        \midrule
        \rowcolor{gray!12} \multicolumn{5}{@{}l}{\textit{2D-based policies}} \\
        \hspace{0.5em}DP-T~\cite{chi2025diffusionpolicy}        & 0.1B   & $38.3_{\pm 2.3}$ & $20.0_{\pm 2.3}$ & $31.0_{\pm 1.7}$ \\
        \hspace{0.5em}$\pi_{0.5}$~\cite{intelligence2025pi05}   & 3.2B & $70.3_{\pm 2.2}$ & $19.5_{\pm 3.3}$ & $50.0_{\pm 1.9}$ \\
        \hspace{0.5em}GR00T N1.6~\cite{bjorck2025gr00t}         & 3B   & $66.8_{\pm 2.3}$ & $43.0_{\pm 2.7}$ & $57.3_{\pm 1.8}$ \\
        \hspace{0.5em}FastWAM~\cite{yuan2026fastwam}            & 6B   & $59.1_{\pm 1.8}$ & $43.5_{\pm 2.4}$ & $52.9_{\pm 1.5}$ \\
        \midrule
        \rowcolor{gray!12} \multicolumn{5}{@{}l}{\textit{3D-based policies}} \\
        \hspace{0.5em}PointACT~\cite{chen2026pointact}          & 0.3B & $43.1_{\pm 2.6}$ & $19.5_{\pm 2.4}$ & $33.7_{\pm 1.8}$ \\
        \hspace{0.5em}PointWAM~\cite{park2026pointwam}          & 1.2B & $82.2_{\pm 1.5}$ & $49.1_{\pm 2.8}$ & $69.0_{\pm 2.0}$ \\
        \rowcolor{lightgreen}
        \hspace{0.5em}\ourmodel{} (Ours)                        & 0.25B & $\textbf{86.2}_{\pm 2.7}$ & $\textbf{64.0}_{\pm 2.4}$ & $\textbf{77.3}_{\pm 1.4}$ \\
        \bottomrule
    \end{tabular}
    \vspace{-1.5em}
\end{table}

\noindent\textbf{DexJoCo.}
To evaluate whether \ourmodel generalizes beyond conventional gripper-based manipulation, we consider DexJoCo~\cite{wang2026dexjoco}, a challenging benchmark featuring high-DoF Allegro dexterous hands.
Its tasks require fine-grained finger control, tool manipulation, and coordinated interactions between two dexterous hands.
Following~\cite{park2026pointwam}, we evaluate on 10 tasks, including 6 single-arm and 4 bimanual tasks, using the provided human demonstrations to train a single multi-task policy for 40k iterations.
We evaluate 50 episodes per task across three random seeds.
As shown in~\Cref{tab:dexjoco_sota_grouped}, \ourmodel achieves 77.3\% average success, outperforming PointWAM~\cite{park2026pointwam} which is pretrained on large-scale human videos.
The improvement is more significant on bimanual dexterous tasks, highlighting the effectiveness of \ourmodel to high-dimensional control spaces.

\begin{table}[t]
\centering
\small
\setlength{\tabcolsep}{3pt}
\caption{Performance on the DexSuite benchmark, grouped by end-effector (number of tasks in parentheses).
}
\label{tab:dexsuite_sota}
\vspace{-0.8em}
\begin{tabular}{lcccc}
\toprule
Method & Size & \makecell{Gripper (5)} & \makecell{Allegro (9)} & \makecell{Avg (14)} \\
\midrule
\rowcolor{gray!12} \multicolumn{5}{l}{\textit{Single-task policies}} \\
\quad ACT~\cite{zhao2023act}                           & 62M & $53.2_{\pm 1.9}$ & $48.9_{\pm 1.5}$ & $50.4_{\pm 1.2}$ \\
\quad DP~\cite{chi2025diffusionpolicy}  & 79M    & $42.8_{\pm 2.5}$ & $44.3_{\pm 2.5}$ & $43.8_{\pm 1.8}$ \\
\midrule
\rowcolor{gray!12} \multicolumn{5}{l}{\textit{Multi-task multi-robot policies}} \\
\quad $\pi_0$~\cite{black2024pi0}                      & 3.1B  & $68.3_{\pm 2.8}$ & $70.1_{\pm 1.4}$ & $69.4_{\pm 0.9}$ \\
\quad  $\pi_{0.5}$~\cite{intelligence2025pi05}          & 3.2B  & $62.7_{\pm 2.3}$ & $64.7_{\pm 1.6}$ & $64.0_{\pm 3.1}$ \\
\rowcolor{lightgreen}
\quad \ourmodel (Ours)                                & 0.25B & $\textbf{87.7}_{\pm 2.4}$ & $\textbf{85.1}_{\pm 2.2}$ & $\textbf{86.0}_{\pm 2.7}$ \\
\bottomrule
\end{tabular}
\vspace{-1.5em}
\end{table}

\noindent\textbf{DexSuite.}
We further evaluate on DexSuite~\cite{houssaini2026dexsuite}, a recent benchmark that covers diverse robot arms, parallel-jaw grippers, and dexterous hands.
Unlike benchmarks with a fixed robot embodiment, DexSuite enables evaluation across different manipulation platforms and action spaces.
We train a unified multi-task policy covering 3 embodiments and 14 task-embodiment pairs with 50 demonstrations each. No embodiment-specific prompts or parameters are introduced. We evaluate 50 episodes per task-embodiment pair across three random seeds, following the official evaluation protocol.
Results in~\Cref{tab:dexsuite_sota} show that our model achieves 86\% average success, demonstrating a unified 3D policy across different robot embodiments and manipulation complexities.

\noindent\textbf{Real-world Experiments.}
For real robot evaluation, we use a 6-DoF UR5 robotic arm with a Weiss WSG50 gripper and UMI-style parallel jaws. 
The robot operates over a 1m $\times$ 1m tabletop workspace observed by a fixed Orbbec Femto Mega RGB-D camera. Observations include an posed RGB-D image and a 7D absolute joint state. Actions are specified as 7D joint state, and we use forward kinematics to obtain the end-effector pose as auxiliary anchor. 
We collect 50 human teleoperation demonstrations for 5 tasks and train a single multi-task policy fine-tuned from the pretrained DROID checkpoint.
We evaluate on 10 test episodes per task, with randomized initial object placements and distractors in each episode.
Further details of the real robot setup are presented in the supplementary material.

\begin{figure}[t]
    \centering
    \includegraphics[width=\linewidth]{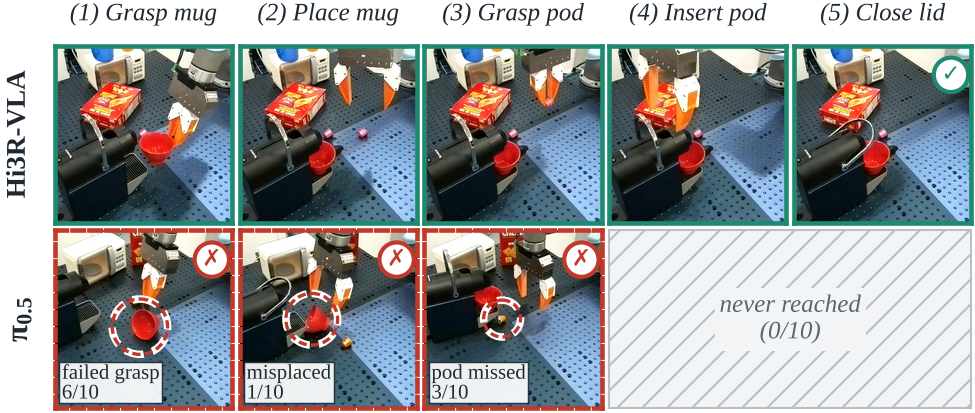}
    \vspace{-1.8em}
    \caption{\textbf{Qualitative comparison on \textit{Prepare coffee}.} $\pi_{0.5}$ failures are shown with their frequency over 10 trials, while \ourmodel completes the full task for 3 out of 10 trials.}
    \label{fig:real_robot_qual}
    \vspace{-1.5em}
\end{figure}

We compare \ourmodel against $\pi_{0.5}$~\cite{intelligence2025pi05} under the same fine-tuning and evaluation protocol.
Results in~\Cref{fig:real_robot_results} demonstrate that \ourmodel can be effectively deployed on physical robots, despite imperfect depth measurements and real-world control uncertainties.
\ourmodel achieves particularly strong performance on tasks requiring precise manipulation like \emph{Prepare coffee}.
\Cref{fig:real_robot_qual} shows qualitative comparisons between \ourmodel and $\pi_{0.5}$.

\section{Conclusion}
\label{sec:conclusion}

We present \ourmodel, a geometry-grounded 3D policy that unifies heterogeneous vision, language, and action representations in a shared 3D space.
It integrates spatially aligned visual and language features into hierarchical point representations, while spatial anchors and 3D positional encoding bridge scene geometry and action generation.
Extensive experiments across diverse robotic manipulation benchmarks demonstrate strong performance with a compact model.
Our ablations highlight the importance of aligning action targets with scene geometry, integrating multi-modalities into 3D representations, and leveraging robot pretraining, providing practical insights into the design of effective and efficient 3D-based manipulation policies.
Limitation and future work are discussed in~\Cref{app:limitation} of the supplementary material.

\section*{Acknowledgments}
{
This work was granted access to HPC resources of IDRIS under the allocation AD011014846R2 and A0211018228 made by GENCI. 
It was funded in part by the French government under management of Agence Nationale de la Recherche as part of the “France 2030" program, reference ANR-23-IACL-0008 (PR[AI]RIE-PSAI projet), the ANR project 3D-GEM (ANR-25-CE23-7777-01), and the Paris Île-de-France Région in the frame of the DIM AI4IDF. 
Cordelia Schmid would like to acknowledge the support by the Körber European Science Prize.
}

\section*{LLM Usage Statement}
We used generative AI tools, primarily GPT and Claude, to assist with code development and implementation, experimental data processing and visualization, literature search, and manuscript polishing. 
All research ideas, methodological designs, experimental protocols, and scientific conclusions were developed by the authors. All AI-assisted code, analyses, and manuscript content were reviewed and verified by the authors. The authors take full responsibility for the final content of this work, including all text, claims, results, and artifacts produced with the assistance of generative AI.

{
    \small
    \bibliographystyle{ieeenat_fullname}
    \bibliography{main}
}

\clearpage
\setcounter{page}{1}
\maketitlesupplementary

\appendix

The supplementary material is organized as follows.
We first present model and implementation details in \Cref{app:implementation}, the DROID dataset and pretraining setup in \Cref{app:droid}, and downstream benchmark setups in \Cref{app:benchmarks}.
We then provide ablation studies on DexJoCo in \Cref{app:dexjoco_ablation}, the effects of pretraining in \Cref{app:pretrain_ablation}, per-task results across benchmarks in \Cref{app:additional_results}, and real-robot experiments in \Cref{app:realworld}.
Finally, we present qualitative results in \Cref{app:qualitative}, and discuss limitations and future work in \Cref{app:limitation}.

\section{Implementation Details}
\label{app:implementation}

\subsection{Model Architecture and Configuration}
\label{app:model_architecture}

We provide additional implementation details of \ourmodel, including the multimodal point encoder and the flow-matching action expert.

\noindent\textbf{Overview.}
Given $V$ calibrated RGB-D views, a proprioceptive state $\mathbf{s}_t$, and a language instruction $\mathcal{L}$, \ourmodel first lifts the views into a point cloud $\mathcal{P}_t$ expressed in a centered metric frame (\Cref{app:point_preprocessing}).
The multimodal point encoder $\mathcal{E}_P$ then encodes $\mathcal{P}_t$ with a hierarchical Point Transformer V3 (PTv3)~\cite{wu2024ptv3} initialized from Utonia~\cite{zhang2026utonia}, injecting multi-view SigLIP2~\cite{tschannen2025siglip2} image features at every encoder stage and conditioning every point block on the instruction (\Cref{app:multimodal_point_encoder}).
Finally, a flow-matching action expert predicts the action chunk $\mathbf{A}_t$ by attending to the multi-scale point representations $\{\mathcal{M}^{\ell}\}_{\ell=1}^{L}$ and to the image and language tokens (\Cref{app:action_expert}), using spatial anchors and 3D RoPE to relate actions and points in metric space (\Cref{app:spatial_encoding}).
The encoders run once per observation, and their outputs are reused for all flow integration steps.
The model configuration is summarized in \Cref{tab:model_config} and the training hyperparameters in \Cref{tab:training_config}.

\noindent\textbf{Point transformer architecture.}
The PTv3 encoder has $L=5$ stages, indexed by $s\in\{0,\dots,4\}$ from fine to coarse, with $(3,3,3,12,3)$ transformer blocks, $(54,108,216,432,576)$ channels, and $(3,6,12,24,32)$ attention heads (head dimension $18$).
Consecutive stages are connected by grid pooling with stride $2$, so the cell size grows from the $1$\,cm input grid to $2$, $4$, $8$, and $16$\,cm.
Grid pooling uses max pooling over the projected features and averages the coordinates of the points merged into each cell.
Each block consists of a sparse convolutional conditional positional encoding, serialized patch self-attention, and an MLP, each with a pre-LayerNorm residual connection.
Patch attention uses a patch size of $1024$ points grouped by one of the four serializations ($z$-order, Hilbert, and their transposed variants), which are randomly permuted during training.
It also applies the 3D RoPE of Utonia with base $10$ and maximum frequency $1$\,rad/m.
The encoder is initialized from the released Utonia checkpoint~\cite{zhang2026utonia}.
Since this checkpoint expects $9$ input channels, we keep the first $6$ columns of its input stem to consume the $(x,y,z,r,g,b)$ values of each point.

We denote the point features of encoder stage $s$ as $\mathbf{h}^{(s)}_i$.
The multi-scale representations start from the encoder bottleneck, $\mathbf{f}^{1}_i=\mathbf{h}^{(4)}_i$, and a lightweight, newly initialized decoder recovers the four finer scales through grid unpooling and skip connections:
\begin{equation}
  \mathbf{f}^{\ell}_i
  = \phi^{\mathrm{skip}}_{\ell}\big(\mathbf{h}^{(L-\ell)}_i\big)
  + \phi^{\mathrm{up}}_{\ell}\big(\mathbf{f}^{\ell-1}_{\pi_{\ell}(i)}\big),
  \qquad \ell \in \{2,\dots,L\},
  \label{eq:supp_unpool}
\end{equation}
where $\pi_{\ell}(i)$ is the parent cell of point $i$ at the coarser scale, and $\phi$ is a Linear-LayerNorm-GELU projection.
The decoder contains no transformer blocks, as adding them yields similar performance but increases training cost in our early experiments.
The resulting representations $\mathcal{M}^{1},\dots,\mathcal{M}^{5}$ have $d_\ell = 576, 432, 216, 108, 108$ channels.
The coordinate $\mathbf{p}^{\ell}_i$ of each point is the centroid of its cell at scale $\ell$.
Before the action expert, each $\mathbf{f}^{\ell}_i$ is mapped to the hidden size $512$ by a scale-specific LayerNorm and linear layer.

\begin{table}[t]
    \centering
    \caption{Model configuration of \ourmodel.}
    \label{tab:model_config}
    \small
    \setlength{\tabcolsep}{4pt}
    \begin{tabular}{@{}ll@{}}
        \toprule
        \textbf{Component} & \textbf{Configuration} \\
        \midrule
        Point input & $\le 4096$ points, $1$\,cm voxels \\
        Point encoder & PTv3, Utonia init. \\
        Point encoder stages $L$ & 5, grid-pooling stride 2 \\
        Encoder blocks per stage & $(3,3,3,12,3)$ \\
        Point feature dim.\ $d_\ell$ & 54--576, projected to 512 \\
        Image encoder & SigLIP2-So400m/16 (frozen) \\
        Text encoder & SigLIP2-So400m (frozen) \\
        Image/text dim.\ $d_V$, $d_T$ & 512 \\
        Point-image fusion & 5 stages \\
        Point-text fusion & AdaLN, 24 blocks \\
        Action expert hidden size & 512 \\
        Action expert blocks & 15 (3 per scale) \\
        Attention heads & 8 (head dimension $d=64$) \\
        Action chunk horizon $H$ & 32 by default \\ 
        RoPE $\omega_{\max}$, $\theta$, $d_r$ & 50\,rad/m, 100, 20 \\
        Flow integration steps & 10 (Euler) \\
        Trainable parameters & $<250$M \\
        \bottomrule
    \end{tabular}
\end{table}

\subsection{Point Cloud Preprocessing}
\label{app:point_preprocessing}

\noindent\textbf{Depth back-projection and coordinate transformation.}
Each view $v$ provides an RGB image $I^{v}_t$, a metric depth map $Z^{v}_t$, intrinsics $\mathbf{K}^{v}$, and a world-to-camera transform $\mathbf{E}^{v}\in SE(3)$, where the extrinsics of wrist cameras are updated at every frame.
The images are first resized to the benchmark resolution (typically $256\times256$), using bilinear interpolation for RGB and nearest-neighbor interpolation for depth, and the intrinsics are updated accordingly.
Photometric augmentation (brightness, contrast, and saturation jitter of $0.2$) is applied before reconstruction, so that the point colors remain consistent with the image features.
A pixel $\mathbf{u}=(u,u')$ with valid depth $z=Z^{v}_t(\mathbf{u})$ is lifted to the world frame as
\begin{equation}
  \begin{bmatrix} \mathbf{p} \\ 1 \end{bmatrix}
  = (\mathbf{E}^{v})^{-1}
  \begin{bmatrix} (\mathbf{K}^{v})^{-1}[zu,\,zu',\,z]^\top \\ 1 \end{bmatrix},
  \label{eq:supp_backproject}
\end{equation}
and inherits the RGB value of the same pixel.
A depth value is valid only if it is finite and lies within a camera-specific range $[z_{\min},z_{\max}]$, and invalid pixels are not lifted.
Cameras without depth, such as the wrist camera of DROID~\cite{khazatsky2024droid}, contribute no points but are still used as image inputs.
The world frame is the robot base frame or the world frame of each benchmark.

\noindent\textbf{Workspace boundaries and voxelization.}
Background regions such as the floor, walls, and distant clutter are removed with a workspace box.
The box is chosen per benchmark to contain the reachable range of the end-effector.
The cropped points are voxelized with a voxel size of $1$\,cm, where points sharing the voxel index $\lfloor \mathbf{p}/0.01 \rfloor$ are merged by averaging their coordinates and colors.

\noindent\textbf{Point sampling and normalization.}
We remove isolated points, which mostly come from flying pixels at depth discontinuities, by discarding points with fewer than $2$ neighbors within a radius of $2.5$\,cm.
This test is skipped if it would leave fewer than $128$ points.
At most $4096$ points are then kept by uniform random sampling.
The points are then centered on their mean $\boldsymbol{\mu}=\frac{1}{N}\sum_{i=1}^{N}\mathbf{p}_i$, which defines the point frame
\begin{equation}
  \tilde{\mathbf{p}} = \mathbf{G}\,\mathbf{p},
  \qquad
  \mathbf{G} =
  \begin{bmatrix} \mathbf{I} & -\boldsymbol{\mu} \\ \mathbf{0}^\top & 1 \end{bmatrix}.
  \label{eq:supp_point_frame}
\end{equation}
Colors are mapped to $[-1,1]$.
The same transform $\mathbf{G}$ is applied to the end-effector positions in the robot state and in absolute actions, while delta actions are unchanged.
As a result, points, states, and actions share one metric coordinate frame.
The camera pose with respect to this frame, $\mathbf{E}^{v}\mathbf{G}^{-1}$, is used to project points back into the images.
The same pipeline is applied at inference.
Rotations in states and actions are converted to the 6D representation, and state and action vectors are normalized to zero mean and unit variance with per-benchmark statistics.

\subsection{Multimodal Point Encoder}
\label{app:multimodal_point_encoder}

\noindent\textbf{Multi-view image feature extraction and projection.}
Each view is letterboxed to the default SigLIP2 resolution of $256\times256$ by an aspect-preserving resize and zero padding, normalized to $[-1,1]$, and encoded by the frozen SigLIP2-So400m/16 vision tower without its pooling head.
This yields a $16\times16$ grid of $1152$-dimensional patch features $\mathbf{U}^{v}_t$ after the final LayerNorm of the tower.
A trainable projection maps them to the feature map
\begin{equation}
  \mathbf{V}^{v}_t
  = \operatorname{LN}_{\mathrm{out}}\big(
    \mathbf{W}_V\,\operatorname{LN}(\mathbf{U}^{v}_t)
  \big) \in \mathbb{R}^{16\times16\times d_V},
  \label{eq:supp_image_proj}
\end{equation}
where $d_V=512$, and the LayerNorms and the linear layer $\mathbf{W}_V$ are applied to each patch.
The vision tower is shared by all views, and the number of views can differ across samples and datasets.

\noindent\textbf{Visibility handling and multi-view feature aggregation.}
For a point $\tilde{\mathbf{p}}_i$ and a view $v$, $\operatorname{proj}_v$ first computes the camera-frame position $\mathbf{E}^{v}\mathbf{G}^{-1}\tilde{\mathbf{p}}_i$ with depth $z^{v}_i$ and its pixel location $\mathbf{u}^{v}_i$ using $\mathbf{K}^{v}$.
It then maps $\mathbf{u}^{v}_i$ to the letterboxed image and to continuous patch-grid coordinates with the pixel-center convention, i.e., a letterboxed pixel coordinate $c$ maps to $(c+0.5)/16-0.5$ along each axis.
The pair is visible, $\nu^{v}_i=1$, if and only if $z^{v}_i>0$, $\mathbf{u}^{v}_i$ falls inside the image, the depth at the nearest pixel is valid, and the point is not occluded:
\begin{equation}
  \big|\, z^{v}_i - Z^{v}_t(\lfloor \mathbf{u}^{v}_i \rceil) \,\big|
  \le \epsilon_{\mathrm{abs}} + \epsilon_{\mathrm{rel}}\, z^{v}_i,
  \label{eq:supp_visibility}
\end{equation}
where $\lfloor\cdot\rceil$ rounds to the nearest pixel, $\epsilon_{\mathrm{abs}}=2$\,cm, and $\epsilon_{\mathrm{rel}}=0.02$.
The nearest-pixel depth lookup avoids mixing foreground and background depth values at object boundaries.
The image feature of each point is the average of the bilinearly sampled features $\mathbf{g}^{v}_i = \operatorname{Bilinear}(\mathbf{V}^{v}_t, \operatorname{proj}_v(\tilde{\mathbf{p}}_i))$ over its visible views:
\begin{equation}
  \mathbf{g}_i
  = \frac{\sum_{v=1}^{V} \nu^{v}_i\, \mathbf{g}^{v}_i}
         {\max\!\big(1, \sum_{v=1}^{V} \nu^{v}_i\big)} \in \mathbb{R}^{d_V}.
  \label{eq:supp_view_mean}
\end{equation}
This average is invariant to the camera order and supports any number of views without view-specific parameters.
Points that are not visible in any view receive $\mathbf{g}_i=\mathbf{0}$, which is rare because every point is reconstructed from one of the views.

\noindent\textbf{Image-point fusion at different PTv3 stages.}
The image features are carried as a separate per-point attribute instead of only being concatenated to the PTv3 input as in~\cite{huang2026pointworld}.
At the input resolution, they are added after the stem:
\begin{equation}
  \mathbf{h}^{(0)}_i
  = \operatorname{Stem}(\tilde{\mathbf{p}}_i, \mathbf{rgb}_i)
  + \mathbf{W}_0\,\operatorname{LN}\big(\mathbf{g}^{(0)}_i\big),
  \label{eq:supp_inject0}
\end{equation}
where $\mathbf{g}^{(0)}_i=\mathbf{g}_i$.
At each grid pooling $s \in \{1,\dots,4\}$, the image features are averaged over the set $\mathcal{C}_s(j)$ of points merged into cell $j$, using the same cluster assignment as PTv3, and added to the pooled point features before the transformer blocks of stage $s$:
\begin{equation}
  \begin{aligned}
    \mathbf{g}^{(s)}_j
    &= \frac{1}{|\mathcal{C}_s(j)|}\sum_{i\in\mathcal{C}_s(j)} \mathbf{g}^{(s-1)}_i, \\
    \mathbf{h}^{(s)}_j
    &= \operatorname{Pool}_s\big(\mathbf{h}^{(s-1)}\big)_j
    + \mathbf{W}_s\,\operatorname{LN}\big(\mathbf{g}^{(s)}_j\big).
  \end{aligned}
  \label{eq:supp_inject}
\end{equation}
We use average rather than max pooling for the image features because they describe the average appearance and semantics of a cell.
Each $\mathbf{W}_s\in\mathbb{R}^{d_s\times d_V}$ is a bias-free linear layer initialized to zero, where $d_s$ is the channel width of stage $s$.
Hence, the point encoder exactly reproduces the pretrained Utonia encoder at initialization, and image information is introduced gradually during training.
We do not inject image features into the decoder, which receives image-conditioned features through the skip connections in \Cref{eq:supp_unpool}.

\noindent\textbf{Language conditioning in PTv3.}
Instructions are lowercased, tokenized, and padded to the $64$-token context of the SigLIP2 text tower, following SigLIP pretraining.
The text transformer is frozen, while its final LayerNorm and a new linear layer that maps $1152$ to $d_T=512$ dimensions are trained.
We keep the tokens up to the end-of-sequence token together with the last context position, which SigLIP uses as its pooled embedding, giving the language embeddings $\mathbf{T}=[\mathbf{w}_1;\dots;\mathbf{w}_{N_T}]\in\mathbb{R}^{N_T\times d_T}$.
Padding tokens are masked in all attention operations.

The global language representation $\mathbf{c}_T$ is obtained by multi-head attentive pooling with $8$ heads, each with its own learnable query $\mathbf{q}_h\in\mathbb{R}^{64}$:
\begin{equation}
  \begin{aligned}
    \alpha_{h,n}
    &= \operatorname*{softmax}_{n}
      \Big(\frac{\mathbf{q}_h^\top \mathbf{k}_{h,n}}{\sqrt{64}}\Big), \\
    \mathbf{c}_T
    &= \mathbf{W}_c
      \Big[\textstyle\sum_n \alpha_{1,n}\mathbf{w}_n;\,\dots;\,
           \textstyle\sum_n \alpha_{8,n}\mathbf{w}_n\Big],
  \end{aligned}
  \label{eq:supp_text_pool}
\end{equation}
where $\mathbf{k}_{h,n}$ is the $h$-th head of $\mathbf{W}_k\operatorname{LN}(\mathbf{w}_n)$, and padding tokens receive zero weight.
Each head pools the full, unprojected token embeddings $\mathbf{w}_n$, and the linear layer $\mathbf{W}_c$ fuses the $8$ concatenated head outputs into $\mathbf{c}_T\in\mathbb{R}^{512}$.
The vector $\mathbf{c}_T$ modulates both pre-normalization layers, before self-attention and before the MLP, of all $24$ encoder blocks:
\begin{equation}
  \begin{aligned}
    \operatorname{AdaLN}(\mathbf{h}_i;\mathbf{c}_T)
    &= (1+\boldsymbol{\gamma})\odot\operatorname{LN}(\mathbf{h}_i)
    + \boldsymbol{\beta}, \\
    [\boldsymbol{\gamma};\boldsymbol{\beta}]
    &= \mathbf{W}_{\mathrm{ada}}\,\operatorname{SiLU}(\mathbf{c}_T),
  \end{aligned}
  \label{eq:supp_adaln}
\end{equation}
where $\operatorname{LN}$ keeps its pretrained affine parameters and $\mathbf{W}_{\mathrm{ada}}$, separate for each normalization layer, is initialized to zero.
Therefore, the pretrained PTv3 is also preserved at initialization.
The same $\mathbf{c}_T$ is shared by all points of a sample and by all stages.

\subsection{Spatial Anchors and 3D RoPE}
\label{app:spatial_encoding}

Each action token is assigned a spatial anchor $\mathbf{x}_i\in\mathbb{R}^3$ in the point frame of \Cref{eq:supp_point_frame}, and each point token at scale $\ell$ uses its coordinate $\mathbf{p}^{\ell}_j$.

\noindent\textbf{Spatial anchors for different action spaces.}
The anchor of the state token of each arm is its current end-effector position.
The anchors of the action tokens depend on the action space.
\begin{itemize}
  \item \textbf{Executable anchors.}
  When actions are absolute end-effector poses, as in RLBench, the first three action dimensions are the target end-effector position.
  This position directly serves as the anchor of the action token.
  For dexterous hands, the anchor is the end-effector position of the arm, and the hand joint targets are additional action dimensions.
  \item \textbf{Auxiliary anchors.}
  When actions are relative end-effector motions, as in RoboCasa, the action itself specifies no position in the scene.
  We therefore prepend to each delta action $\Delta\mathbf{a}_{t+j}$ the absolute end-effector position $\mathbf{y}_{t+j+1}$ of the next state, which is the position reached after executing the action:
  \begin{equation}
    \hat{\mathbf{a}}_{t+j} = \big[\,\mathbf{y}_{t+j+1};\;\Delta\mathbf{a}_{t+j}\,\big].
    \label{eq:supp_aux_anchor}
  \end{equation}
  At the end of an episode, the last position is repeated.
  The prefix is predicted jointly with the delta action by the same flow-matching objective, serves only as the anchor, and is discarded at execution time.
\end{itemize}
In both cases, the anchor of an action token is read from the first three dimensions of its noisy action.
The anchors are thus computed from the current sample $\mathbf{A}^{\tau}_t$ rather than from ground truth, which makes them available at inference.
During integration, they move from noise toward the predicted trajectory as $\tau\rightarrow1$.

\noindent\textbf{Coordinate normalization.}
The flow-matching inputs and targets are normalized, whereas the anchors must be metric.
We therefore denormalize only the positions used for 3D RoPE:
\begin{equation}
  \begin{aligned}
    \mathbf{x}_{k,j} &= \boldsymbol{\sigma}_a\odot\big[\mathbf{a}^{\tau}_{t+j,k}\big]_{1:3} + \boldsymbol{\mu}_a, \\
    \mathbf{x}_{k,\mathrm{s}} &= \boldsymbol{\sigma}_s\odot\big[\mathbf{s}_{t,k}\big]_{1:3} + \boldsymbol{\mu}_s,
  \end{aligned}
  \label{eq:supp_denorm}
\end{equation}
where $(\boldsymbol{\mu}_a,\boldsymbol{\sigma}_a)$ and $(\boldsymbol{\mu}_s,\boldsymbol{\sigma}_s)$ are the per-benchmark statistics of the action and state positions, respectively.
Auxiliary anchors use the state statistics.
Since points, states, and actions are expressed in the same centered frame, the offsets between anchors and points are physical displacements in meters, which are consistent across datasets.

\noindent\textbf{3D RoPE formulation and frequency parameters.}
With head dimension $d=64$, we use $d_r=20$ rotary dimensions per axis, i.e., $d_r/2=10$ feature pairs per axis.
The remaining $d-3d_r=4$ dimensions are not rotated.
The rotation of a position $\mathbf{x}=(x_1,x_2,x_3)$ is block diagonal,
\begin{equation}
  R(\mathbf{x})
  = \bigoplus_{a=1}^{3}\,\bigoplus_{m=0}^{d_r/2-1} R_m(x_a)
  \;\oplus\; \mathbf{I}_{d-3d_r},
  \label{eq:supp_rope_rot}
\end{equation}
where $R_m$ is the $2\times2$ rotation with frequency $\omega_m = \omega_{\max}\theta^{-2m/d_r}$ defined in the main paper, $\omega_{\max}=50$\,rad/m, and $\theta=100$.
These values give frequencies from $50$ to $0.79$\,rad/m, i.e., wavelengths $2\pi/\omega_m$ from $12.6$\,cm to $7.9$\,m.
The maximum frequency sets the finest resolved offset, as a displacement of $2$\,cm, twice the voxel size, rotates the fastest feature pair by one radian.
The base is chosen such that the slowest frequency $\omega_{\min}=\omega_{\max}\theta^{-(d_r-2)/d_r}$ does not wrap around within the workspace.
Its phase remains unambiguous for offsets up to $\pi/\omega_{\min}\approx4$\,m, which covers workspaces of about $2$\,m.
Since all rotations in \Cref{eq:supp_rope_rot} commute and are orthogonal, the attention score depends only on the relative position, $(R(\mathbf{x}_i)\mathbf{q}_i)^\top(R(\mathbf{p}_j)\mathbf{k}_j)=\mathbf{q}_i^\top R(\mathbf{p}_j-\mathbf{x}_i)\mathbf{k}_j$.

\noindent\textbf{Application to attention.}
3D RoPE is applied in two attention layers of every action expert block.
\begin{itemize}
  \item \textbf{Action self-attention.}
  The queries and keys of the state and action tokens are both rotated by their anchors, so the interaction among actions and robot states, depends on their relative displacement.
  The temporal order of the action tokens is encoded separately by the trajectory-index embedding and is not used in RoPE.
  \item \textbf{Action-to-point cross-attention.}
  The queries are rotated by the action anchors and the keys by the point coordinates $\mathbf{p}^{\ell(b)}_j$ of the scale read by block $b$.
  Each action token thus attends to points according to their offset from its intended end-effector position.
  The values are not rotated.
\end{itemize}
The language and image tokens have no 3D position, so their cross-attention uses no positional encoding.
The rotated point keys depend only on the observation and are computed once, whereas the query rotations are recomputed at every integration step from the updated anchors.

\subsection{Flow-Matching Action Expert}
\label{app:action_expert}

\noindent\textbf{Action representation.}
For a robot with $K\le2$ arms, each action $\mathbf{a}_{t+j}$ in the chunk $\mathbf{A}_t$ contains one vector per arm, consisting of the end-effector position, a 6D rotation, and the gripper or dexterous-hand joint targets.
To train different embodiments with one model, action and state vectors are zero-padded to $32$ and $64$ dimensions, respectively.
Padded dimensions, padded timesteps beyond the episode end, and padded arms are excluded from the loss and from attention.

\noindent\textbf{Action tokens.}
For each arm $k$, the action expert operates on one state token followed by $H$ action tokens:
\begin{equation}
  \begin{aligned}
    \mathbf{z}_{k,\mathrm{s}} &= \mathbf{W}_s\mathbf{s}_{t,k} + \mathbf{e}^{\mathrm{arm}}_k + \mathbf{c}_\tau, \\
    \mathbf{z}_{k,j} &= \mathbf{W}_a\mathbf{a}^{\tau}_{t+j,k}
      + \mathbf{e}^{\mathrm{traj}}_j + \mathbf{e}^{\mathrm{arm}}_k + \mathbf{c}_\tau,
  \end{aligned}
  \label{eq:supp_tokens}
\end{equation}
where $j \in \{0,\dots,H-1\}$, $\mathbf{a}^{\tau}_{t+j,k}$ is the noisy action of arm $k$ in $\mathbf{A}^{\tau}_t$, $\mathbf{e}^{\mathrm{traj}}_j$ is a learned trajectory-index embedding, and $\mathbf{e}^{\mathrm{arm}}_k$ is a learned arm embedding used only for bimanual robots.
The flow-time embedding $\mathbf{c}_\tau$ is computed by a Linear--SiLU--Linear MLP from sinusoidal features of $\tau$, following $\pi_0$~\cite{black2024pi0}.
The token sequences of all arms are concatenated so that the arms attend to each other.

\noindent\textbf{Blocks.}
Each of the $15$ blocks updates the action tokens $\mathbf{Z}$, while the point, image, and language features remain fixed:
\begin{align}
  \mathbf{Z} &\leftarrow \mathbf{Z}
    + \operatorname{SelfAttn}\big(\operatorname{LN}(\mathbf{Z})\big), \\
  \mathbf{Z} &\leftarrow \mathbf{Z}
    + \operatorname{CrossAttn}\big(\operatorname{LN}(\mathbf{Z}),\, \operatorname{LN}([\mathbf{T};\mathbf{V}_t])\big), \\
  \mathbf{Z} &\leftarrow \mathbf{Z}
    + \operatorname{CrossAttn}\big(\operatorname{LN}(\mathbf{Z}),\, \operatorname{LN}(\mathcal{M}^{\ell(b)})\big), \\
  \mathbf{Z} &\leftarrow \mathbf{Z}
    + \operatorname{FFN}\big((1+\boldsymbol{\gamma}_\tau)\odot\operatorname{LN}(\mathbf{Z}) + \boldsymbol{\beta}_\tau\big),
\end{align}
where $\mathcal{M}^{\ell(b)}$ is the projected point representation read by block $b$.
The FFN is modulated by the flow time with $[\boldsymbol{\gamma}_\tau;\boldsymbol{\beta}_\tau] = \mathbf{W}_\tau\mathbf{c}_\tau$, where $\mathbf{W}_\tau$ is initialized to zero.
The language and image tokens share one cross-attention, while the point tokens use a separate cross-attention module.
The self-attention and the point cross-attention use 3D RoPE (\Cref{app:spatial_encoding}).
The FFN has an expansion ratio of $4$, and dropout of $0.1$ is applied in attention and FFN layers.
The blocks attend to the point scales in contiguous groups of three, from $\mathcal{M}^{1}$ in blocks $1$--$3$ to $\mathcal{M}^{5}$ in blocks $13$--$15$.
We do not concatenate all scales into one memory, since the finest scale would dominate the number of tokens.
After the last block, the state tokens are discarded, and a linear head predicts the velocity of each action token.
The keys and values of the language, image, and point tokens are computed once per observation and reused for all integration steps.

\begin{table}[t]
    \centering
    \caption{Training hyperparameters of \ourmodel.}
    \label{tab:training_config}
    \small
    \setlength{\tabcolsep}{4pt}
    \begin{tabular}{@{}ll@{}}
        \toprule
        \textbf{Hyperparameter} & \textbf{Value} \\
        \midrule
        Optimizer & AdamW, $\beta=(0.9, 0.999)$ \\
        Learning rate & $10^{-4}$ \\
        Weight decay & $0.05$ \\
        Learning rate schedule & $3\%$ warmup, cosine to $10^{-8}$ \\
        Gradient clipping & global norm $3.0$ \\
        Global batch size & 128 ($2\times$ H100) \\
        Training iterations & 40k--100k \\ %
        Mixed precision & bf16 (action expert in fp32) \\
        Dropout & 0.1 \\
        Flow time distribution & $\tau=1-u$, $u\sim\operatorname{Beta}(1.5,1)$ \\
        Points per sample & $\le 4096$ \\
        Color jitter & $0.2$ \\
        \bottomrule
    \end{tabular}
\end{table}

\noindent\textbf{Flow-matching objective.}
The flow matching loss is computed on normalized actions and averaged over valid entries.
With the predicted velocity $\hat{\mathbf{U}}=v_\theta(\mathbf{A}_t^\tau,\tau,\mathcal{O}_t)$ and the target velocity $\mathbf{U}=\mathbf{A}_t-\boldsymbol{\epsilon}$, the loss is
\begin{equation}
  \mathcal{L}_{\mathrm{FM}}
  = \frac{1}{|\Omega|}\sum_{(j,k)\in\Omega}\;\sum_{c\in\mathcal{D}_k}
    \big( \hat{U}_{j,k,c} - U_{j,k,c} \big)^2,
  \label{eq:supp_fm_loss}
\end{equation}
where $\Omega$ is the set of valid timestep-arm pairs and $\mathcal{D}_k$ is the set of valid action dimensions of arm $k$.
The flow time is sampled as $\tau=1-u$ with $u\sim\operatorname{Beta}(1.5,1)$, which places more samples in the noise-dominated part of the path.
At inference, we start from $\mathbf{A}_t^{0}\sim\mathcal{N}(\mathbf{0},\mathbf{I})$ and integrate with $10$ Euler steps, $\mathbf{A}_t^{\tau+\Delta} = \mathbf{A}_t^{\tau} + \Delta\, v_\theta(\mathbf{A}_t^{\tau},\tau,\mathcal{O}_t)$ with $\Delta=0.1$.
Padded action dimensions are reset to zero after every step.

\begin{figure*}[t]
\centering
\includegraphics[width=\linewidth]{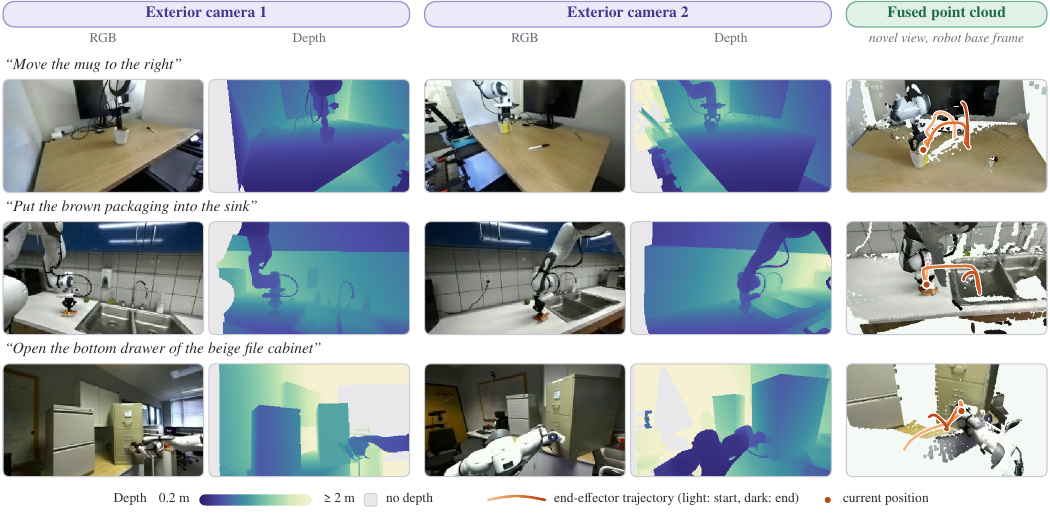}
\caption{Samples of the DROID pretraining data. For each demonstration, we show
the RGB image and depth map of the two exterior cameras, and the point cloud
obtained by back-projecting both views into the robot base frame, rendered from
a novel viewpoint. The orange curve is the end-effector trajectory of the
episode, from light (start) to dark (end), and the dot is the current
end-effector position.}
\label{fig:supp_droid_samples}
\end{figure*}

\subsection{Training Details}
\label{app:training}

\Cref{tab:training_config} lists the training hyperparameters.
We optimize all trainable parameters with AdamW using a single learning rate.
Weight decay is not applied to biases and normalization parameters.
The learning rate is linearly warmed up over the first $3\%$ of the iterations and then follows a cosine decay.
We use bf16 mixed precision for the frozen SigLIP2 encoders and for the linear and attention layers of PTv3, whose sparse convolutions run in float32.
The point projections, the action expert, and the loss are computed in float32, except for the attention kernels, which use bf16 inputs.
Models are trained with a global batch size of $128$ on $2$ NVIDIA H100 GPUs for $40$k to $100$k iterations depending on the benchmark, which takes less than $1$ second per iteration.
Hence, it takes around 8 hours to 1 day for each experiment on downstream benchmarks.

\section{DROID Pretraining}
\label{app:droid}

We use real-world
demonstrations from DROID~\cite{khazatsky2024droid} dataset to pretrain \ourmodel.

\noindent\textbf{Sources and selection.}
DROID~\cite{khazatsky2024droid} is released without depth maps and with noisy
camera extrinsics. We take the per-frame stereo depth, estimated with
FoundationStereo~\cite{wen2025foundationstereo}, and the optimized extrinsics of
the two exterior cameras from PointWorld~\cite{huang2026pointworld}, and the RGB
videos, robot states, commanded actions and language instructions from
RoboInter~\cite{li2026robointer}. Of the 76.5k successful DROID demonstrations,
35{,}811 have PointWorld depth and calibration. We discard 573 of them whose
instruction does not describe a task (\eg, ``no action''), which leaves 35{,}238
demonstrations (Table~\ref{tab:supp_droid_stats}).
Figure~\ref{fig:supp_droid_samples} shows some examples in the cleaned dataset.

\noindent\textbf{Observations and actions.}
Each frame provides two exterior RGB-D views at $320{\times}180$ resolution with
their intrinsics and extrinsics, and a wrist RGB view without depth. The
extrinsics map the robot base frame to each camera, so point clouds, states and
actions share the base frame. At training time, we back-project the pixels of
both exterior views whose depth lies in $[0.2, 3.0]$\,m, which gives close to
100k points per frame before downsampling. The state is the 8-D end-effector
pose, \ie position, quaternion and gripper opening. The action is the
end-effector pose commanded by the teleoperation controller. We keep the native
15\,Hz rate of DROID.

\begin{table}[t]
\centering
\small
\caption{Statistics of the DROID pretraining dataset.}
\label{tab:supp_droid_stats}
\begin{tabular}{ll}
\toprule
Demonstrations & 35{,}238 \\
Frames & 10.1M (187 hours at 15\,Hz) \\
Episode length & median 14.7\,s \\
Distinct instructions & 25{,}387 \\
Cameras & 2 exterior RGB-D, 1 wrist RGB \\
State / action & 8-D end-effector pose and gripper \\
\bottomrule
\end{tabular}
\end{table}

\noindent\textbf{Pretraining configuration.}
DROID pretraining follows the training details in~\Cref{app:training}, except that we use $4$ NVIDIA H100 GPUs with a global batch size of $256$ and train for $2$ epochs ($\sim 79$k iterations).

\begin{table*}[t]
    \centering
    \caption{Per-benchmark configurations.
    All cameras provide RGB-D: Images are $256\times256$, except for RoboTwin 2.0 ($240\times320$).
    We convert the actions of each benchmark to absolute targets in the world frame, with the end-effector (EEF) pose given as position and 6D rotation, so that the EEF positions serve as executable spatial anchors except on RoboTwin 2.0.}
    \label{tab:benchmark_configs}
    \small
    \setlength{\tabcolsep}{4pt}
    \newcolumntype{L}[1]{>{\raggedright\arraybackslash}p{#1}}
    \begin{tabular}{@{}lL{3.1cm}L{2.5cm}L{2.6cm}L{2.5cm}L{2.9cm}@{}}
        \toprule
        \textbf{Benchmark} & \textbf{Embodiment} & \textbf{Cameras} & \textbf{Action per arm} & \textbf{Training data} & \textbf{Evaluation} \\
        \midrule
        RLBench~\cite{james2020rlbench} & Franka Panda, parallel gripper & front & next keyframe EEF pose, gripper & 10 tasks $\times$ 100 demos & 3 seeds $\times$ 100 episodes per task \\
        \addlinespace
        RoboCasa~\cite{nasiriany2024robocasa} & Franka Panda on an Omron mobile base & left / right agentview, wrist & EEF pose, gripper, mobile base & 24 tasks $\times$ 50 human demos & 3 seeds $\times$ 50 episodes per task \\
        \addlinespace
        RoboTwin 2.0~\cite{chen2025robotwin2} & Aloha-AgileX, two 6-DoF arms with grippers & head, left / right wrist & EEF pose (auxiliary anchor), gripper, arm joints & 50 tasks $\times$ 50 clean demos & 100 episodes per task, clean and randomized \\
        \addlinespace
        \multirow{2}{*}{DexJoCo~\cite{wang2026dexjoco}} & Franka Panda, 16-DoF Allegro hand & front, right ego-view, wrist & EEF pose, 16-DoF hand joints & 6 tasks $\times$ 100 demos & 3 seeds $\times$ 50 episodes per task \\
         & two Franka Panda arms, two Allegro hands & ego-view, left / right wrist & EEF pose, 16-DoF hand joints & 4 bimanual tasks $\times$ 100 demos & 3 seeds $\times$ 50 episodes per task \\
        \addlinespace
        DexSuite~\cite{houssaini2026dexsuite} & Franka Panda, Allegro hand, UMI or Robotiq gripper & front, wrist & EEF pose, hand joints or gripper & 14 task-embodiment pairs $\times$ 50 demos & 3 seeds $\times$ 50 episodes per task \\
        \bottomrule
    \end{tabular}
\end{table*}

\section{Robotic Manipulation Benchmarks}
\label{app:benchmarks}

\Cref{tab:benchmark_configs} summarizes the robot embodiments, observations, action spaces, training data, and evaluation protocols of the simulation benchmarks.
All benchmarks are converted to a common RGB-D format, from which point clouds are reconstructed online as described in \Cref{app:point_preprocessing}.
Demonstrations whose replay in the simulator fails to reach the task success are discarded.
For simplicity, the table reports the original number of demonstrations before filtering.

\subsection{RoboCasa24}
\label{app:robocasa}

\noindent\textbf{Benchmark and tasks.}
RoboCasa~\cite{nasiriany2024robocasa} is a large-scale simulation framework of household kitchens built on robosuite and MuJoCo.
We use its $24$ atomic tasks, which cover pick-and-place between the counter, cabinets, sink, microwave, and stove ($8$ tasks), opening and closing single and double doors ($4$) and drawers ($2$), turning the sink faucet on or off and turning the spout ($3$), turning the stove on or off ($2$), the coffee machine ($3$), and the microwave buttons ($2$).
The robot is a Franka Panda arm mounted on an Omron mobile base and controlled by an operational-space controller at $20$\,Hz.

\noindent\textbf{Demonstrations.}
We train on the original human demonstrations, using $50$ demonstrations per task and $1{,}200$ in total.
Since the released demonstrations contain no depth, we replay every demonstration in the simulator and re-render RGB-D observations at $256\times256$.
The $73$ demonstrations whose replay does not succeed are discarded, leaving $1{,}127$ demonstrations.
One policy is trained on all $24$ tasks.

\noindent\textbf{Observations and actions.}
The policy observes the left and right agent-view cameras and the eye-in-hand camera.
The original actions are delta commands of the end-effector pose, together with gripper, mobile-base, torso, and mode commands, expressed in the frame of the mobile base.
We convert each delta command to an absolute end-effector goal by applying the scaled translation and rotation to the end-effector pose before the action, and keep the other commands unchanged.
At evaluation time, the predicted absolute goals are converted back to delta commands relative to the current end-effector pose, and are executed by the original controller.

\noindent\textbf{Evaluation.}
Following Cosmos Policy~\cite{kim2026cosmospolicy}, each task is evaluated on $50$ episodes, where every block of $10$ episodes uses one of five kitchen layout and style combinations, and we report the mean success rate over $3$ random seeds.
The episode horizon follows the task, ranging from $300$ to $1{,}000$ control steps, and an episode succeeds once the task success condition is met.
The policy predicts chunks of $32$ actions and executes the first $16$ before replanning.

\subsection{RLBench}
\label{app:rlbench}

\noindent\textbf{Tasks and demonstrations.}
RLBench~\cite{james2020rlbench} is a simulation benchmark built on CoppeliaSim, in which a Franka Panda arm with a parallel gripper performs tabletop tasks.
We follow the $10$-task setting~\cite{liu2025hybridvla,chen2026pointact}, which contains close box, close laptop lid, toilet seat down, sweep to dustpan, close fridge, phone on base, take umbrella out of umbrella stand, take frame off hanger, stack wine, and water plants. %
Each task is trained with $100$ demonstrations, and one policy is trained on all tasks.

\noindent\textbf{Keyframes.}
As is common for RLBench, the policy predicts keyframe actions instead of dense trajectories.
A frame is a keyframe if the gripper opens or closes, if the joint velocities are close to zero, or if it is the last frame of the demonstration.
Each training sample pairs the observation at a keyframe with the end-effector pose and gripper state of the next keyframes. 

\noindent\textbf{Observations and actions.}
We only use the front RGB-D camera.
At evaluation time, each predicted pose is reached with a motion planner, followed by the gripper action.

\noindent\textbf{Evaluation.}
Each task is evaluated on $100$ episodes for $3$ times with newly sampled scene initializations.
An episode succeeds if the task is completed within $25$ keyframe steps, and failures of the motion planner are counted as failed episodes.
We use a chunk size of $H=2$ and replan at each keyframe.

\subsection{DexJoCo}
\label{app:dexjoco}

\noindent\textbf{Benchmark and embodiment.}
DexJoCo~\cite{wang2026dexjoco} is a MuJoCo benchmark of task-oriented dexterous manipulation, in which each arm is a Franka Panda equipped with a $16$-DoF Allegro hand.
We use $6$ single-arm tasks, namely click mouse, fold glasses, hammer nail, pick bucket, pinch tongs, and water plant, and $4$ bimanual tasks, namely assembly, Hanoi, microwave cooking, and photograph.
We follow PointWAM~\cite{park2026pointwam} and exclude the bimanual unlock-iPad task, since $42$ of its $100$ demonstrations fail to replay.

\noindent\textbf{Demonstrations.}
Each task provides $100$ teleoperated demonstrations.
We replay each demonstration in the simulator to render RGB-D observations at $256\times256$, and discard the demonstrations whose replay fails.
One policy is trained on all single-arm and bimanual tasks, and the two arms of bimanual tasks are distinguished by the learned arm embedding (\Cref{app:action_expert}).

\noindent\textbf{Observations and actions.}
Single-arm tasks use the front, right ego-view, and wrist cameras, and bimanual tasks use the ego-view and the two wrist cameras.
The action of each arm consists of the absolute end-effector pose and the $16$ absolute joint targets of the hand, which are native to DexJoCo, and is expressed in the frame of the robot base.
The predicted actions are directly executed by the simulator controllers.

\noindent\textbf{Evaluation.}
Each task is evaluated on $50$ episodes with randomized object placements, and we report the mean success rate over $3$ random seeds.
The episode horizon ranges from $1{,}000$ to $1{,}500$ steps depending on the task.
The policy predicts chunks of $32$ actions and executes the first $16$ before replanning.

\subsection{RoboTwin 2.0}
\label{app:robotwin}

\noindent\textbf{Benchmark and tasks.}
RoboTwin 2.0~\cite{chen2025robotwin2} is a SAPIEN benchmark of $50$ bimanual tabletop tasks performed by the Aloha-AgileX robot, which has two $6$-DoF arms with parallel grippers.

\noindent\textbf{Demonstrations.}
We train one policy on the official clean demonstrations, with $50$ demonstrations per task and $2{,}500$ in total.
Since the released data contain no depth, we rebuild each scene from its seed, replay the recorded joint and gripper targets, and render RGB-D observations at the native resolution of $240\times320$.
The $92$ demonstrations whose replay does not succeed are discarded, leaving $2{,}408$ demonstrations.

\noindent\textbf{Observations and actions.}
The policy observes the static head camera and the two wrist cameras.
The action of each arm contains the absolute end-effector pose as the auxiliary anchor, the gripper opening, and the $6$ joint targets of the next frame, expressed in the simulator world frame.
At evaluation time, the predicted joint and gripper targets are executed by the joint-position controller of the benchmark.

\noindent\textbf{Evaluation.}
Following the official protocol, each task is evaluated on $100$ episodes in two settings.
The clean setting matches the training distribution with seen instructions, while the randomized setting adds cluttered distractors, random background textures, lighting changes, and table-height variations, together with unseen instructions.
Evaluation seeds for which the expert planner fails are skipped, and the episode horizon follows the task-specific step limit of the benchmark.
The policy predicts chunks of $32$ actions and executes the first $16$ before replanning.

\section{Ablation Studies on DexJoCo}
\label{app:dexjoco_ablation}

In this section, we extend the ablation studies on RoboCasa24 in~\Cref{sec:ablation_expr} to DexJoCo.
Unlike RoboCasa24, which focuses on single-arm manipulation with a parallel gripper, DexJoCo includes both single-arm and bimanual dexterous manipulation with Allegro hands.
However, DexJoCo exhibits less variation in evaluation, as the scene layouts and object instances remain fixed across episodes.

\begin{figure*}[t]
  \centering
  \begin{subfigure}[t]{0.3364\textwidth}
    \captionsetup{margin={0.1635\linewidth,0pt}}
    \includegraphics[width=\linewidth]{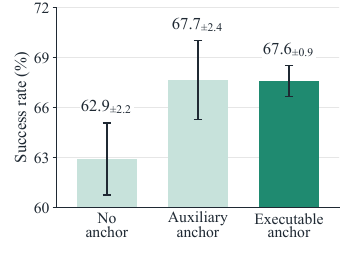}
    \caption{Spatial anchor.}
    \label{fig:dexjoco_ablation_anchor}
  \end{subfigure}%
  \begin{subfigure}[t]{0.3802\textwidth}
    \includegraphics[width=\linewidth]{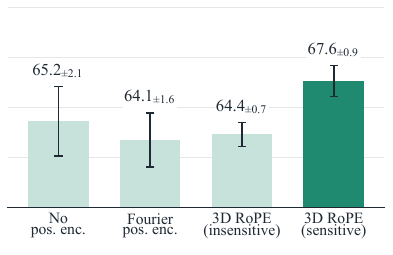}
    \caption{3D position encoding.}
    \label{fig:dexjoco_ablation_posenc}
  \end{subfigure}%
  \begin{subfigure}[t]{0.2834\textwidth}
    \includegraphics[width=\linewidth]{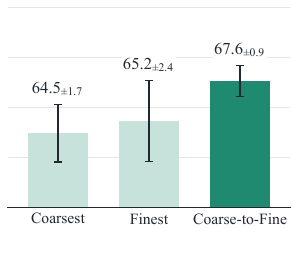}
    \caption{Point features.}
    \label{fig:dexjoco_ablation_features}
  \end{subfigure}
  \caption{Ablation of the geometry-aware action expert on DexJoCo. Results are reported as mean success rates over three seeds.}
  \label{fig:dexjoco_ablation}
\end{figure*}

\subsection{How to Ground Actions in 3D Geometry?}

To isolate the contribution of 3D geometry, we remove the image encoder and condition the action expert solely on point cloud and language features in this subsection.

\noindent\textbf{Spatial anchors.}
We compare three action representations: \emph{no anchor}, \emph{auxiliary anchor}, and \emph{executable anchor}.
Without an anchor, the action expert predicts delta end-effector (EEF) poses and hand joint commands without action-side 3D positional encoding.
The auxiliary anchor additionally introduces an absolute EEF pose as a prefix to the action sequence for spatial encoding, while retaining delta EEF pose and hand joints for execution.
The executable anchor directly predicts the absolute EEF pose as part of the executable action, alongside hand joint commands.

As shown in~\Cref{fig:dexjoco_ablation_anchor}, both spatial anchors improve success rates by more than 3 points over the no-anchor baseline.
The auxiliary and executable anchors achieve comparable averaged performance, although the executable anchor exhibits lower variance.
This may be because dexterous manipulation relies more heavily on hand joint control, reducing the differences between the two anchor representations.
Nevertheless, both variants demonstrate the benefits of explicitly grounding actions in 3D geometry.

\noindent\textbf{3D positional encoding.}
We investigate how different positional encoding strategies affect spatial interactions between action queries and scene points, comparing \emph{no positional encoding}, \emph{Fourier features}, \emph{3D RoPE (insensitive)}, and \emph{3D RoPE (sensitive)}.
As shown in~\Cref{fig:dexjoco_ablation_posenc}, 3D RoPE outperforms both Fourier features and the no-encoding baseline.
The sensitive variant performs better than the insensitive variant, highlighting the importance of selecting appropriate RoPE frequencies to capture fine-grained spatial relationships.

\noindent\textbf{Multi-scale point features.}
We compare conditioning the action expert on a single feature scale with a coarse-to-fine strategy that progressively incorporates finer-scale features across decoder blocks.
As shown in~\Cref{fig:dexjoco_ablation_features}, fine-scale features outperform coarse-scale features, while coarse-to-fine conditioning achieves the best performance, confirming the benefits of progressively refining spatial information during action generation.

All these ablations reveal trends consistent with RoboCasa24, demonstrating that our geometry-aware action expert generalizes across substantially different robot embodiments and manipulation tasks.

\subsection{How to Integrate Vision, Language and 3D?}

We further investigate how to integrate visual and language features into the 3D point encoder and action expert.

\begin{table}[t]
  \centering
  \begin{tabular}{cccc}
    \toprule
    Point-text & Point-image & Action-image & \multirow{2}{*}{SR (\%)} \\
    fusion     & fusion      & condition    & \\
    \midrule
    $\times$   & $\times$    & $\times$     & 68.2$_{\pm 4.2}$ \\
    Cross-attn & $\times$    & $\times$     & 68.3$_{\pm 2.7}$ \\
    AdaLN      & $\times$    & $\times$     & 67.6$_{\pm 0.9}$ \\
    \midrule
    \multirow{4}{*}{AdaLN} & Input & $\times$ & 68.4$_{\pm 1.9}$ \\
               & All stages  & $\times$     & 69.8$_{\pm 0.6}$ \\
               & $\times$    & $\checkmark$ & 72.1$_{\pm 1.6}$ \\
               & All stages  & $\checkmark$ & \textbf{75.1}$_{\pm 2.5}$ \\
    \bottomrule
  \end{tabular}
  \caption{Ablation of multimodal fusion in point encoder and conditioning in action expert on DexJoCo.}
  \label{tab:dexjoco_fusion_ablation}
\end{table}

\noindent\textbf{Point-text fusion.}
In~\Cref{tab:dexjoco_fusion_ablation} (upper block), we compare three strategies for incorporating language into point representations: no fusion, token-level cross-attention, and pooled text conditioning via AdaLN.
Unlike RoboCasa24, DexJoCo contains only 10 tasks with distinct objects across tasks, making language conditioning less critical for distinguishing task-relevant objects.
Hence, the point-text fusion has a negligible effect on performance.

\noindent\textbf{Image-point fusion.}
We examine whether and where to incorporate spatially aligned image features into the point encoder.
As shown in~\Cref{tab:dexjoco_fusion_ablation}, multi-stage fusion outperforms both input-only fusion and no fusion, consistent with the results on RoboCasa24.
This supports the benefits of integrating visual appearance with explicit 3D geometry.

\noindent\textbf{Action-level image conditioning.}
We further evaluate direct image conditioning in the action expert in~\Cref{tab:dexjoco_fusion_ablation}.
Unlike RoboCasa24, direct image conditioning provides larger gains on DexJoCo, while remaining complementary to image-point fusion.

\section{Ablation: How Does Pretraining Help?}
\label{app:pretrain_ablation}

\begin{figure*}[t]
  \centering
  \begin{subfigure}[t]{0.32\textwidth}
    \includegraphics[width=\linewidth]{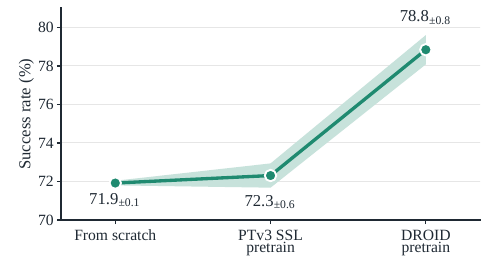}
    \caption{Pretraining effect.}
    \label{fig:pretrain_sr}
  \end{subfigure}
  \hfill
  \begin{subfigure}[t]{0.32\textwidth}
    \includegraphics[width=\linewidth]{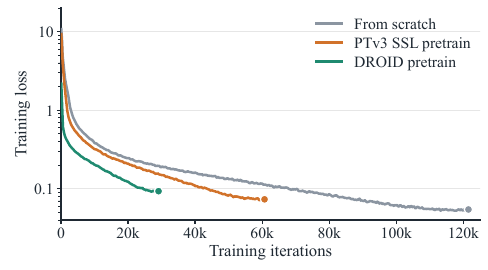}
    \caption{Training loss.}
    \label{fig:pretrain_loss}
  \end{subfigure}
  \hfill
  \begin{subfigure}[t]{0.32\textwidth}
    \includegraphics[width=\linewidth]{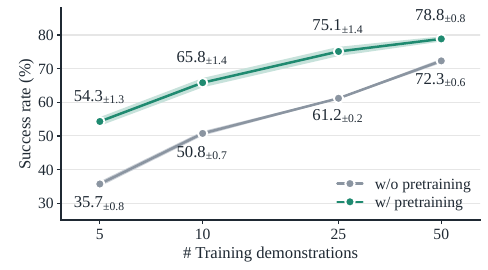}
    \caption{Data efficiency.}
    \label{fig:data_efficiency}
  \end{subfigure}
  \caption{Effect of pretraining for \ourmodel on RoboCasa24. 
    (a) Mean success rate of \ourmodel  without pretraining, with only pretrained point encoder, and pretrained on DROID dataset.
    (b) Training loss against iterations; each dot marks the end of a run. (c) Performance of using different demonstrations w/ and w/o pretraining on DROID.}
  \label{fig:pretrain_ablation}
  \vspace{-1em}
\end{figure*}

We study how pretraining affects policy performance and learning efficiency.
Unless otherwise specified, all following ablations are conducted on RoboCasa24.

\noindent\textbf{Pretraining strategies.}
We compare training from scratch, initializing the point encoder with self-supervised PTv3 pretraining~\cite{zhang2026utonia}, and further pretraining the policy on real-robot dataset DROID~\cite{khazatsky2024droid}.
Given their different optimization speeds, we use different training budgets: 50, 25, 12 epochs, corresponding to approximately 120k, 60k, and 30k iterations, respectively.
We report the final performance of each model and compare their training dynamics in~\Cref{fig:pretrain_sr} and~\ref{fig:pretrain_loss}.
As shown in~\Cref{fig:pretrain_loss}, the pretrained point encoder achieves faster optimization than training from scratch.
However, the improvement in success rate is modest, from 71.9\% to 72.3\% (\Cref{fig:pretrain_sr}).
This suggests that generic geometric pretraining alone provides limited benefits for downstream action prediction.
DROID pretraining substantially improves the success rate to 78.8\%, outperforming PTv3 self-supervised pretraining by 6.5 points.
It also achieves this improvement with substantially fewer fine-tuning iterations, indicating both faster adaptation and better downstream performance.

\begin{table}[t]
    \centering
    \caption{
        Effect of DROID pretraining across robotic manipulation benchmarks.
        RoboTwin2 results are reported as clean$\rightarrow$clean /
        clean$\rightarrow$randomized, where the latter evaluates
        generalization to unseen scenes.
        Gains are measured in percentage points (pp).
    }
    \label{tab:pretrain_effects_benchmarks}
    \begin{tabular}{lccc}
        \toprule
        Benchmark & Pretrain & SR (\%) & Gain (pp) \\
        \midrule
        \multirow{2}{*}{RoboCasa24}
        & $\times$ & $72.3_{\pm 0.6}$ & -- \\
        & $\checkmark$ & $\mathbf{78.8_{\pm 0.8}}$ & $\textcolor{red}{+6.5}$ \\
        \midrule
        \multirow{2}{*}{RoboTwin2}
        & $\times$ & $86.2$ / 33.4 & -- \\
        & $\checkmark$ & $\mathbf{90.1}$ / $\mathbf{50.5}$ & \textcolor{red}{$+3.9$ / $+17.1$} \\
        \midrule
        \multirow{2}{*}{DexJoCo}
        & $\times$ & $75.1_{\pm 2.5}$ & -- \\
        & $\checkmark$ & $\mathbf{77.3_{\pm 1.4}}$ & \textcolor{red}{$+2.2$} \\
        \bottomrule
    \end{tabular}
\end{table}

\noindent\textbf{Effect of DROID pretraining across benchmarks.}
As shown in~\Cref{tab:pretrain_effects_benchmarks}, DROID pretraining consistently improves performance across all three benchmarks. 
The gains on standard evaluation settings are largest on RoboCasa24, followed by RoboTwin2 (clean) and DexJoCo, potentially reflecting differences in embodiment similarity between the pretraining and downstream datasets.
RoboCasa24 shares the same robot embodiment as DROID, while RoboTwin2 uses different robotic arms but is still gripper-based manipulation, and DexJoCo employs dexterous robotic hands.
Moreover, the gains are much more significant on RoboTwin2 under randomized conditions, with an improvement of 17.1 points. This suggests that pretraining on diverse real-world robot data is especially beneficial for generalization to unseen scenarios, while providing more moderate gains in clean environments where training and evaluation distributions are more closely aligned.

\noindent\textbf{Data efficiency after pretraining.}
We evaluate the effect of DROID pretraining on RoboCasa24 using 5, 10, 25, and 50 demonstrations per task.
Pretraining consistently improves success rates, with larger gains when downstream data are scarce: 18.6 points with 5 demonstrations, compared with 6.5 points with 50 demonstrations.
In addition, the pretrained model achieves 54.3\% success with only 5 demonstrations per task, exceeding the 50.8\% achieved without pretraining using twice as many demonstrations.
These results show that DROID pretraining reduces the amount of task-specific demonstration data needed for effective adaptation.

\section{Additional Quantitative Results}
\label{app:additional_results}

\subsection{Per-Task Results on RoboCasa24}
\label{app:robocasa_results}

\begin{table}[t]
\centering
\small
\setlength{\tabcolsep}{4pt}
\caption{Per-task success rates (\%) on RoboCasa (24 kitchen tasks, 50 demonstrations per task).
\ourmodel reports mean and standard deviation over 3 seeds with 50 episodes each.
Baseline numbers are taken from~\cite{gonzalez2026sawam}. Best results are in \textbf{bold}.}
\label{tab:robocasa24_sota_per_task}
\begin{tabular}{lccc}
\toprule
Task & \makecell{Cosmos-\\Policy\cite{kim2026cosmospolicy}} & \makecell{SA-WAM\\\cite{gonzalez2026sawam}} & \makecell{\ourmodel\\(Ours)} \\
\midrule
\multicolumn{4}{l}{\textit{Pick-and-place (8)}} \\
PnPCounterToCab & $55.3$ & $\textbf{68.0}$ & $64.0_{\pm 10.0}$ \\
PnPCabToCounter & $30.0$ & $50.0$ & $\textbf{66.0}_{\pm 6.9}$ \\
PnPCounterToSink & $58.3$ & $\textbf{79.0}$ & $74.0_{\pm 12.0}$ \\
PnPSinkToCounter & $75.3$ & $\textbf{85.3}$ & $79.3_{\pm 2.3}$ \\
PnPCounterToMicrowave & $35.3$ & $\textbf{44.0}$ & $38.0_{\pm 7.2}$ \\
PnPMicrowaveToCounter & $31.3$ & $\textbf{56.0}$ & $54.0_{\pm 2.0}$ \\
PnPCounterToStove & $61.3$ & $\textbf{84.7}$ & $82.0_{\pm 2.0}$ \\
PnPStoveToCounter & $67.3$ & $\textbf{81.0}$ & $76.7_{\pm 3.1}$ \\
\midrule
\multicolumn{4}{l}{\textit{Doors / drawers (6)}} \\
OpenSingleDoor & $82.7$ & $74.7$ & $\textbf{94.7}_{\pm 1.1}$ \\
CloseSingleDoor & $95.3$ & $97.0$ & $\textbf{100.0}_{\pm 0.0}$ \\
OpenDoubleDoor & $91.3$ & $99.3$ & $\textbf{100.0}_{\pm 0.0}$ \\
CloseDoubleDoor & $88.0$ & $94.0$ & $\textbf{97.3}_{\pm 1.1}$ \\
OpenDrawer & $\textbf{90.7}$ & $\textbf{90.7}$ & $88.0_{\pm 5.3}$ \\
CloseDrawer & $98.7$ & $\textbf{99.3}$ & $\textbf{99.3}_{\pm 1.1}$ \\
\midrule
\multicolumn{4}{l}{\textit{Turn / toggle (7)}} \\
TurnOnStove & $54.3$ & $62.0$ & $\textbf{74.7}_{\pm 4.2}$ \\
TurnOffStove & $20.7$ & $\textbf{28.7}$ & $24.0_{\pm 2.0}$ \\
TurnOnSinkFaucet & $62.7$ & $87.0$ & $\textbf{88.7}_{\pm 3.1}$ \\
TurnOffSinkFaucet & $76.0$ & $\textbf{95.3}$ & $89.3_{\pm 6.4}$ \\
TurnSinkSpout & $76.0$ & $77.3$ & $\textbf{80.0}_{\pm 4.0}$ \\
TurnOnMicrowave & $75.3$ & $92.0$ & $\textbf{96.0}_{\pm 3.5}$ \\
TurnOffMicrowave & $99.3$ & $\textbf{100.0}$ & $99.3_{\pm 1.1}$ \\
\midrule
\multicolumn{4}{l}{\textit{Coffee (3)}} \\
CoffeeSetupMug & $34.7$ & $30.0$ & $\textbf{54.7}_{\pm 4.6}$ \\
CoffeeServeMug & $52.7$ & $65.3$ & $\textbf{76.7}_{\pm 3.1}$ \\
CoffeePressButton & $97.3$ & $\textbf{98.7}$ & $95.3_{\pm 5.0}$ \\
\midrule
\multicolumn{4}{l}{\textit{Category averages}} \\
Pick-and-place (8) & $51.8$ & $\textbf{68.5}$ & $66.8_{\pm 2.4}$ \\
Doors / drawers (6) & $91.1$ & $92.5$ & $\textbf{96.6}_{\pm 0.9}$ \\
Turn / toggle (7) & $66.3$ & $77.5$ & $\textbf{78.9}_{\pm 1.4}$ \\
Coffee (3) & $61.6$ & $64.7$ & $\textbf{75.6}_{\pm 2.5}$ \\
\rowcolor{lightgreen}
Avg (24) & $67.1$ & $76.6$ & $\textbf{78.8}_{\pm 0.8}$ \\
\bottomrule
\end{tabular}
\vspace{-1em}
\end{table}

\Cref{tab:robocasa24_sota_per_task} reports per-task results on the 24 RoboCasa kitchen tasks.
\ourmodel achieves the highest average success rate.
The gains are most pronounced on tasks that require precise object manipulation such as door/drawer and coffee categories.
However, there is still room for improvement, especially on pick-and-place tasks.
These tasks involve a large variety of object instances and categories. 
We expect larger-scale pretraining to benefit such cases.

\subsection{Per-Task Results on RLBench}
\label{app:rlbench_results}

\begin{table*}[t]
    \centering
    \caption{
        Per-task performance on the RLBench-10Task benchmark.
        Size denotes the trainable parameters.
    }
    \label{tab:rlbench_10_sota}
    \resizebox{\textwidth}{!}{%
    \begin{tabular}{@{}lcccccccccccc@{}}
        \toprule
        Model & Size
        & \shortstack{Close\\box}
        & \shortstack{Close\\laptop lid}
        & \shortstack{Toilet\\seat down}
        & \shortstack{Sweep\\to dustpan}
        & \shortstack{Close\\fridge}
        & \shortstack{Phone\\on base}
        & \shortstack{Umbrella\\out}
        & \shortstack{Frame\\off hanger}
        & \shortstack{Wine at\\rack}
        & \shortstack{Water\\plants}
        & Mean \\
        \midrule
        \multicolumn{13}{@{}l}{\textit{2D-based policies}} \\
        ManipLLM~\cite{li2024manipllm}    & 7B   & 50 & 80 & 40 & 20 & 80 & 35 & 10 & 25 & 15 & 20 & 38 \\
        OpenVLA~\cite{kim2024openvla}     & 7B   & 65 & 40 & 75 & 60 & 80 & 20 & 35 & 15 & 10 & 10 & 41 \\
        $\pi_0$~\cite{black2024pi0}     & 3.1B & 90 & 60 & 100 & 30 & 90 & 25 & 35 & 75 & 5 & 45 & 55 \\
        CogACT~\cite{li2024cogact}      & 7B   & 80 & 85 & 90 & 65 & 90 & 50 & 60 & 35 & 25 & 25 & 60 \\
        HybridVLA~\cite{liu2025hybridvla}   & 7B   & 85 & 95 & 100 & 90 & 100 & 50 & 50 & 70 & 50 & 50 & 74 \\
        EO1~\cite{qu2025eo1}         & 3B   & 97 & 99 & 100 & 95 & 83 & 46 & 76 & 40 & 61 & 35 & 73.2 \\
        $\pi_{0.5}$~\cite{intelligence2025pi05} & 3.2B
        & 98.3$_{\pm 0.6}$ & 96.0$_{\pm 2.0}$ & 98.0$_{\pm 2.0}$ & 85.3$_{\pm 0.6}$ & 91.7$_{\pm 2.1}$
        & 67.0$_{\pm 2.0}$ & 83.3$_{\pm 0.6}$ & 49.7$_{\pm 5.5}$ & 69.7$_{\pm 4.6}$ & 35.3$_{\pm 7.1}$
        & 77.4$_{\pm 0.4}$ \\
        \midrule
        \multicolumn{13}{@{}l}{\textit{3D-based policies}} \\
        ACT3D~\cite{gervet2023act3d}       & 1.7M & 94 & 62 & 99 & 80 & 87 & 53 & 97 & 31 & 31 & 11 & 64.5 \\
        PointACT~\cite{chen2026pointact} & 0.3B & 91 & 99 & 96 & 59 & 81 & 99 & 99 & 69 & 90 & 40 & 82.3 \\
        \ourmodel (Ours)   & 0.25B
        & 96.0$_{\pm 2.0}$ & 98.7$_{\pm 0.6}$ & 98.3$_{\pm 1.5}$ & 86.7$_{\pm 6.1}$ & 98.0$_{\pm 1.0}$
        & 96.0$_{\pm 1.0}$ & 98.7$_{\pm 0.6}$ & 65.0$_{\pm 4.6}$ & 95.0$_{\pm 2.6}$ & 74.0$_{\pm 3.0}$
        & \textbf{90.6}$_{\pm 0.3}$ \\
        \bottomrule
    \end{tabular}%
    }
\end{table*}

\Cref{tab:rlbench_10_sota} compares \ourmodel with state-of-the-art 2D- and 3D-based policies on each RLBench 10 tasks.
The largest gains come from tasks that require precise 3D placement or contact-rich motion such as \textit{water plants} and \textit{wine at rack}.
Its main weakness is the \textit{frame off hanger} task, where the hanging frame is easily movable upon contact, making manipulation particularly challenging.

\subsection{Per-Task Results on DexJoCo}
\label{app:dexjoco_results}

\begin{table*}[t]
    \centering
    \caption{
        Performance on the DexJoCo benchmark for dexterous manipulation.
        A single policy is trained across all ten tasks.
        We report mean success rates (\%) with standard deviations over three seeds.
        /B denotes bimanual tasks.
        All baseline results are taken from~\cite{park2026pointwam}.
    }
    \label{tab:dexjoco_sota_per_task}
    \resizebox{\textwidth}{!}{%
    \begin{tabular}{@{}lccccccc@{}}
        \toprule
        \multirow{2}{*}{Task}
        & \multicolumn{4}{c}{2D-based policies}
        & \multicolumn{3}{c}{3D-based policies} \\
        \cmidrule(lr){2-5} \cmidrule(lr){6-8}
        & DP-T~\cite{chi2025diffusionpolicy}
        & $\pi_{0.5}$~\cite{intelligence2025pi05}
        & GR00T N1.6~\cite{bjorck2025gr00t}
        & FastWAM~\cite{yuan2026fastwam}
        & PointACT~\cite{chen2026pointact}
        & PointWAM~\cite{park2026pointwam}
        & \ourmodel{} (Ours) \\
        \midrule
        hammer nail
        & $58.7_{\pm 5.0}$ & $86.7_{\pm 3.1}$ & $82.0_{\pm 6.0}$ & $43.3_{\pm 1.9}$ & $16.0_{\pm 8.0}$ & $85.3_{\pm 2.3}$ & $92.7_{\pm 3.1}$ \\
        click mouse
        & $38.7_{\pm 3.1}$ & $80.7_{\pm 3.1}$ & $67.3_{\pm 7.6}$ & $77.3_{\pm 3.4}$ & $37.3_{\pm 3.1}$ & $97.3_{\pm 2.3}$ & $92.7_{\pm 4.2}$ \\
        pick bucket
        & $55.3_{\pm 7.6}$ & $83.3_{\pm 8.1}$ & $88.7_{\pm 3.1}$ & $81.3_{\pm 6.6}$ & $74.7_{\pm 7.6}$ & $98.0_{\pm 2.0}$ & $94.7_{\pm 3.1}$ \\
        pinch tongs
        & $6.0_{\pm 5.3}$ & $45.3_{\pm 6.1}$ & $28.0_{\pm 4.0}$ & $47.3_{\pm 6.6}$ & $26.7_{\pm 4.2}$ & $52.0_{\pm 5.3}$ & $72.7_{\pm 12.1}$ \\
        fold glasses
        & $11.3_{\pm 5.0}$ & $42.0_{\pm 6.0}$ & $52.0_{\pm 8.0}$ & $62.7_{\pm 3.8}$ & $20.7_{\pm 7.6}$ & $74.0_{\pm 5.3}$ & $76.0_{\pm 9.2}$ \\
        water plant
        & $60.0_{\pm 6.9}$ & $84.0_{\pm 4.0}$ & $82.7_{\pm 3.1}$ & $42.7_{\pm 0.9}$ & $83.3_{\pm 6.4}$ & $86.7_{\pm 3.1}$ & $88.7_{\pm 1.1}$ \\
        hanoi /B
        & $8.0_{\pm 2.0}$ & $6.0_{\pm 0.0}$ & $34.0_{\pm 5.3}$ & $42.0_{\pm 5.9}$ & $10.0_{\pm 2.0}$ & $50.0_{\pm 9.2}$ & $65.3_{\pm 6.4}$ \\
        assembly /B
        & $1.3_{\pm 2.3}$ & $3.3_{\pm 2.3}$ & $6.0_{\pm 4.0}$ & $2.7_{\pm 1.9}$ & $2.0_{\pm 0.0}$ & $9.3_{\pm 4.6}$ & $34.7_{\pm 2.3}$ \\
        microwave /B
        & $42.7_{\pm 6.4}$ & $39.3_{\pm 13.0}$ & $78.7_{\pm 3.1}$ & $88.7_{\pm 4.1}$ & $54.0_{\pm 8.7}$ & $84.0_{\pm 4.0}$ & $83.3_{\pm 3.1}$ \\
        photograph /B
        & $28.0_{\pm 6.0}$ & $29.3_{\pm 1.2}$ & $53.3_{\pm 8.1}$ & $40.7_{\pm 6.2}$ & $12.0_{\pm 3.5}$ & $53.3_{\pm 1.2}$ & $72.7_{\pm 5.8}$ \\
        \midrule
        avg.
        & $31.0_{\pm 1.7}$ & $50.0_{\pm 1.9}$ & $57.3_{\pm 1.8}$ & $52.9_{\pm 1.5}$ & $33.7_{\pm 1.8}$ & $69.0_{\pm 2.0}$ & $\textbf{77.3}_{\pm 1.4}$ \\
        \bottomrule
    \end{tabular}%
    }
\end{table*}

As shown in Table~\ref{tab:dexjoco_sota_per_task}, \ourmodel achieves the best average success rate of 77.3\%, outperforming the state-of-the-art method PointWAM by 8.3 points with fewer parameters.
The improvement, however, is not uniform across embodiments: on single-arm tasks, where strong baselines already approach saturation, \ourmodel gains a modest 4 points on average, whereas on bimanual tasks the gap widens to around 15 points.
The largest gains appear on tasks that demand precise contact and tight coordination, such as assembly, hanoi, photograph and pinch tongs, suggesting that \ourmodel is particularly beneficial when fine-grained spatial reasoning is required.

\subsection{Per-Task Results on RoboTwin2}
\label{app:robotwin_results}

\begin{table*}[t]
\centering
\footnotesize
\setlength{\tabcolsep}{3pt}
\caption{Per-task success rates (\%) on RoboTwin2 (100 trials per task). All models are trained on 50 clean demonstrations per task. Baseline results are from the RoboTwin2 leaderboard. Best result per task and setting in bold.}
\label{tab:robotwin2_pertask}
\begin{tabular}{lcccccc@{\hskip 10pt}lcccccc}
\toprule
\multirow{2}{*}{Task} & \multicolumn{3}{c}{clean$\rightarrow$clean} & \multicolumn{3}{c}{clean$\rightarrow$randomized} & \multirow{2}{*}{Task} & \multicolumn{3}{c}{clean$\rightarrow$clean} & \multicolumn{3}{c}{clean$\rightarrow$randomized} \\
\cmidrule(lr){2-4}\cmidrule(lr){5-7}\cmidrule(lr){9-11}\cmidrule(lr){12-14}
 & \makecell{$\pi_{0.5}$\\\cite{intelligence2025pi05}} & \makecell{4D-WAM\\\cite{yang20264dwam}} & \makecell{Hi3R\\(Ours)} & \makecell{$\pi_{0.5}$\\\cite{intelligence2025pi05}} & \makecell{4D-WAM\\\cite{yang20264dwam}} & \makecell{Hi3R\\(Ours)} &  & \makecell{$\pi_{0.5}$\\\cite{intelligence2025pi05}} & \makecell{4D-WAM\\\cite{yang20264dwam}} & \makecell{Hi3R\\(Ours)} & \makecell{$\pi_{0.5}$\\\cite{intelligence2025pi05}} & \makecell{4D-WAM\\\cite{yang20264dwam}} & \makecell{Hi3R\\(Ours)} \\
\midrule
adjust\_bottle & \textbf{99} & 80 & 98 & \textbf{77} & 36 & 34 & place\_can\_basket & 56 & 83 & \textbf{84} & 10 & 51 & \textbf{55} \\
beat\_block\_hammer & \textbf{93} & 92 & 90 & 23 & 28 & \textbf{41} & place\_cans\_plasticbox & 31 & 94 & \textbf{97} & 73 & 48 & \textbf{83} \\
blocks\_ranking\_rgb & 72 & 88 & \textbf{99} & 44 & 10 & \textbf{70} & place\_container\_plate & 92 & 93 & \textbf{99} & 72 & 82 & \textbf{92} \\
blocks\_ranking\_size & 45 & 54 & \textbf{83} & 21 & 8 & \textbf{49} & place\_dual\_shoes & 70 & 66 & \textbf{84} & \textbf{45} & 19 & 43 \\
click\_alarmclock & 65 & \textbf{100} & \textbf{100} & 50 & 70 & \textbf{85} & place\_empty\_cup & 96 & 99 & \textbf{100} & 80 & 70 & \textbf{83} \\
click\_bell & 31 & 98 & \textbf{100} & 35 & 77 & \textbf{92} & place\_fan & 66 & 48 & \textbf{96} & 26 & 17 & \textbf{34} \\
dump\_bin\_bigbin & 92 & 86 & \textbf{95} & \textbf{84} & 57 & 51 & place\_mouse\_pad & 34 & 54 & \textbf{73} & \textbf{31} & 9 & 23 \\
grab\_roller & 99 & \textbf{100} & \textbf{100} & \textbf{97} & 68 & 77 & place\_object\_basket & 70 & 75 & \textbf{91} & 29 & 40 & \textbf{57} \\
handover\_block & 30 & 79 & \textbf{98} & 12 & \textbf{25} & 20 & place\_object\_scale & 78 & 79 & \textbf{97} & 38 & 24 & \textbf{44} \\
handover\_mic & \textbf{98} & 85 & \textbf{98} & 6 & \textbf{13} & 12 & place\_object\_stand & 85 & 87 & \textbf{98} & 47 & 42 & \textbf{86} \\
hanging\_mug & 17 & \textbf{70} & 49 & 14 & 10 & \textbf{19} & place\_phone\_stand & 65 & 84 & \textbf{92} & 37 & 21 & \textbf{56} \\
lift\_pot & \textbf{98} & 88 & \textbf{98} & \textbf{35} & 9 & 2 & place\_shoe & 82 & 83 & \textbf{98} & 50 & 52 & \textbf{78} \\
move\_can\_pot & 60 & \textbf{94} & 92 & 10 & 17 & \textbf{37} & press\_stapler & 79 & 76 & \textbf{90} & \textbf{63} & 61 & 43 \\
move\_pillbottle\_pad & 46 & 89 & \textbf{100} & \textbf{32} & 22 & 30 & put\_bottles\_dustbin & 78 & 68 & \textbf{88} & \textbf{50} & 19 & 19 \\
move\_playingcard\_away & 84 & 82 & \textbf{98} & \textbf{65} & 35 & 43 & put\_object\_cabinet & 39 & \textbf{69} & 45 & 22 & \textbf{47} & 35 \\
move\_stapler\_pad & 20 & 43 & \textbf{85} & \textbf{18} & 12 & 12 & rotate\_qrcode & 90 & 79 & \textbf{95} & 21 & \textbf{67} & 37 \\
open\_laptop & 93 & 85 & \textbf{95} & 68 & \textbf{82} & 53 & scan\_object & 49 & 73 & \textbf{80} & 40 & \textbf{53} & 21 \\
open\_microwave & 86 & 68 & \textbf{89} & \textbf{59} & 12 & 27 & shake\_bottle & \textbf{100} & 98 & \textbf{100} & \textbf{100} & 72 & 82 \\
pick\_diverse\_bottles & 66 & 77 & \textbf{89} & 42 & 44 & \textbf{50} & shake\_bottle\_horizontally & 99 & 98 & \textbf{100} & \textbf{100} & 70 & 87 \\
pick\_dual\_bottles & 72 & 92 & \textbf{94} & 57 & \textbf{67} & 49 & stack\_blocks\_three & 77 & \textbf{95} & 82 & 25 & \textbf{82} & 31 \\
place\_a2b\_left & 71 & 71 & \textbf{81} & \textbf{49} & 19 & 35 & stack\_blocks\_two & 91 & \textbf{100} & \textbf{100} & 43 & 34 & \textbf{88} \\
place\_a2b\_right & 68 & 62 & \textbf{69} & 39 & \textbf{42} & 37 & stack\_bowls\_three & 81 & 82 & \textbf{93} & 52 & 27 & \textbf{60} \\
place\_bread\_basket & 63 & 87 & \textbf{92} & 60 & \textbf{77} & 76 & stack\_bowls\_two & 94 & 94 & \textbf{99} & 70 & \textbf{81} & 80 \\
place\_bread\_skillet & 76 & 82 & \textbf{91} & 47 & \textbf{69} & 47 & stamp\_seal & 53 & \textbf{74} & 71 & 21 & 10 & \textbf{30} \\
place\_burger\_fries & 82 & 86 & \textbf{98} & 86 & 72 & \textbf{98} & turn\_switch & 52 & \textbf{86} & 71 & 25 & 11 & \textbf{30} \\
\midrule
\multicolumn{7}{c}{} & \textit{Average} & 70.7 & 81.5 & \textbf{90.1} & 46.0 & 41.8 & \textbf{50.5} \\
\bottomrule
\end{tabular}
\end{table*}

\Cref{tab:robotwin2_pertask} breaks down the results over the 50 RoboTwin2 tasks. 
In the clean$\rightarrow$clean setting, \ourmodel achieves the best or tied-best success rate on 42 of 50 tasks. The largest gains appear on tasks that demand precise spatial reasoning about object pose and placement, such as \textit{move\_stapler\_pad} (+42 over the best baseline), \textit{place\_fan} (+30) and \textit{blocks\_ranking\_size} (+29), suggesting that explicit 3D representations help resolve fine-grained geometric relations that are ambiguous from 2D features alone. Under clean$\rightarrow$randomized evaluation, \ourmodel still obtains the highest average and its advantage is most pronounced on placement and stacking tasks, e.g., \textit{stack\_blocks\_two} (+45), \textit{place\_object\_stand} (+39) and \textit{blocks\_ranking\_rgb} (+26), where geometric cues remain informative despite changes in background, lighting and clutter. However, the gain is less uniform than in the clean setting: \ourmodel is best on 23 tasks and falls behind on several contact-rich or articulated-object tasks such as \textit{lift\_pot} and \textit{open\_microwave}. Since all models are fine-tuned only on clean demonstrations, this indicates that out-of-distribution distractors and visual shifts remain challenging, and we expect training with randomized data or stronger pretraining to further close this gap.

\section{Real-World Experiments}
\label{app:realworld}

\begin{center}
\begin{minipage}{\linewidth}
\centering
\expandafter\def\csname @captype\endcsname{figure}
\includegraphics[width=\linewidth]{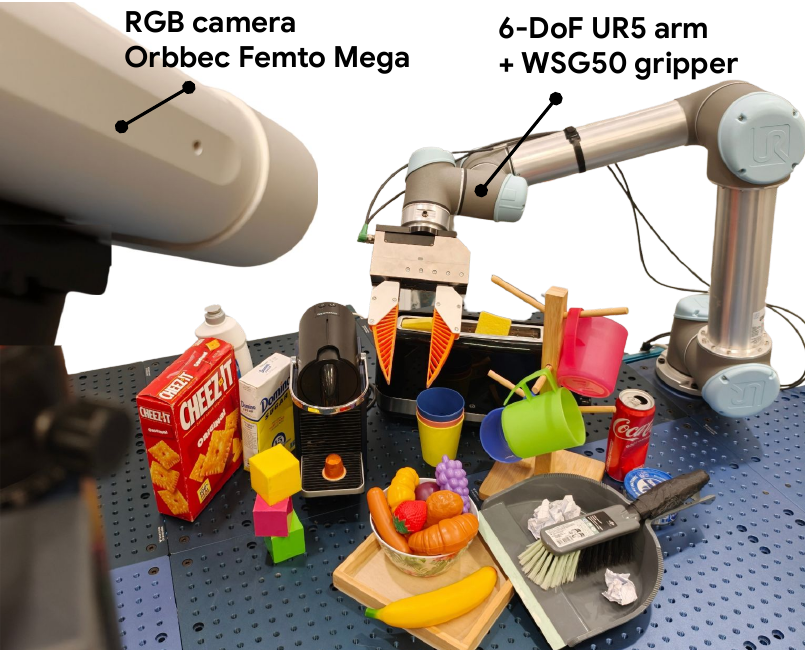}
\caption{UR5 arm with a WSG50 gripper, observed by a fixed Orbbec Femto Mega RGB-D camera, with the task objects and distractors.}
\label{fig:UR5_Setup}
\end{minipage}
\end{center}

\subsection{Robot Setup}
\label{app:realworld_setup}

We use a 6-DoF UR5 robotic arm with a Weiss WSG50 gripper and UMI-style parallel jaws.
The action loop runs at 15 Hz.
The robot operates over a 1\,m${\times}$1\,m tabletop workspace observed by a fixed Orbbec Femto Mega RGB-D camera, see \Cref{fig:UR5_Setup}.
Observations include a single-view RGB-D image and robot proprioceptive state (6-DoF robot joint angles and 1D gripper openness).
Actions are represented in the same 7D joint space.

\begin{figure*}[t]
\centering
\includegraphics[width=\linewidth]{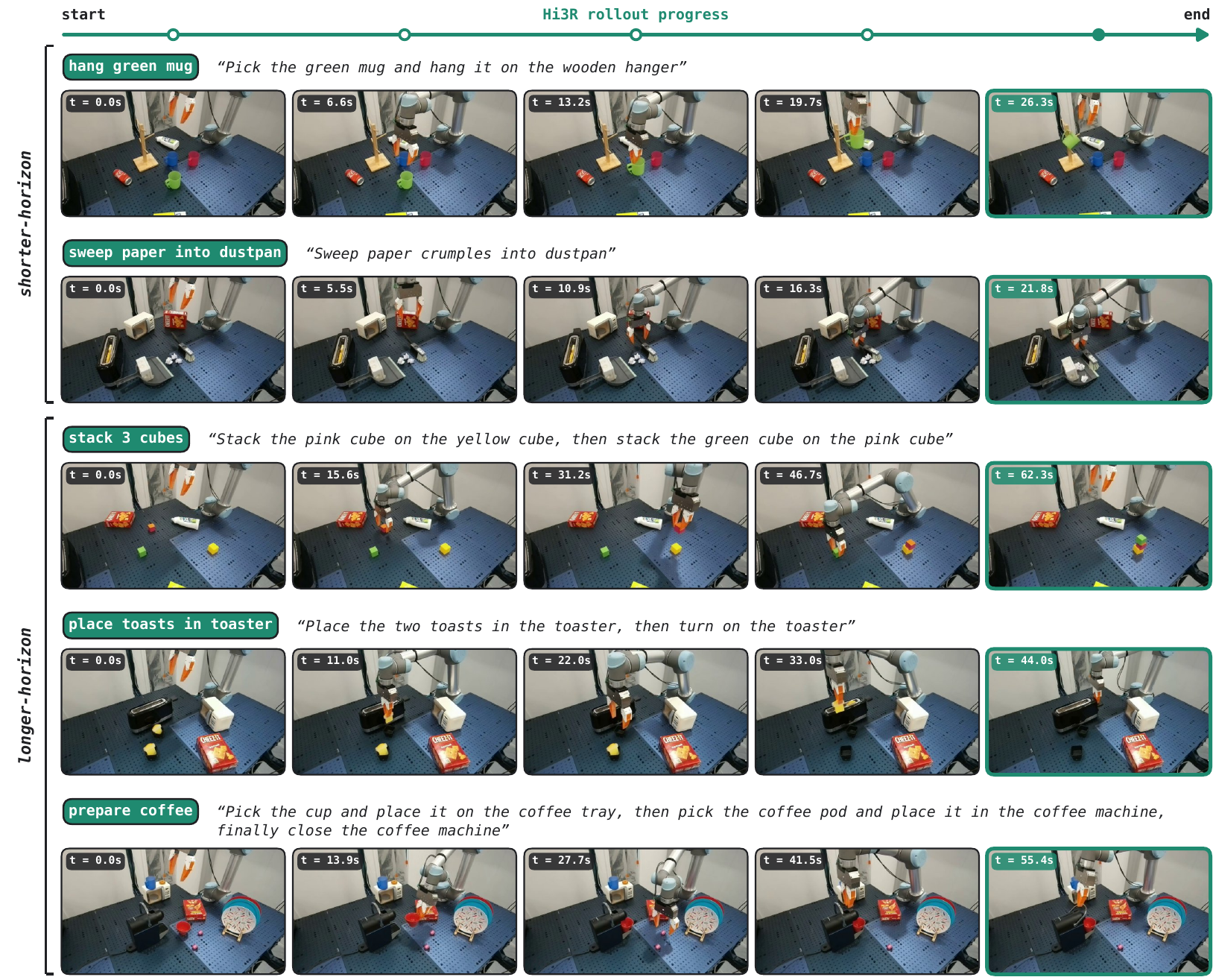}
\caption{\textbf{Qualitative success examples.} Successful rollouts of Hi3R on our UR5 robotic arm across the five
manipulation tasks.}
\label{fig:UR5_successful_rollouts}
\end{figure*}

\begin{figure*}[t]
\centering
\includegraphics[width=\linewidth]{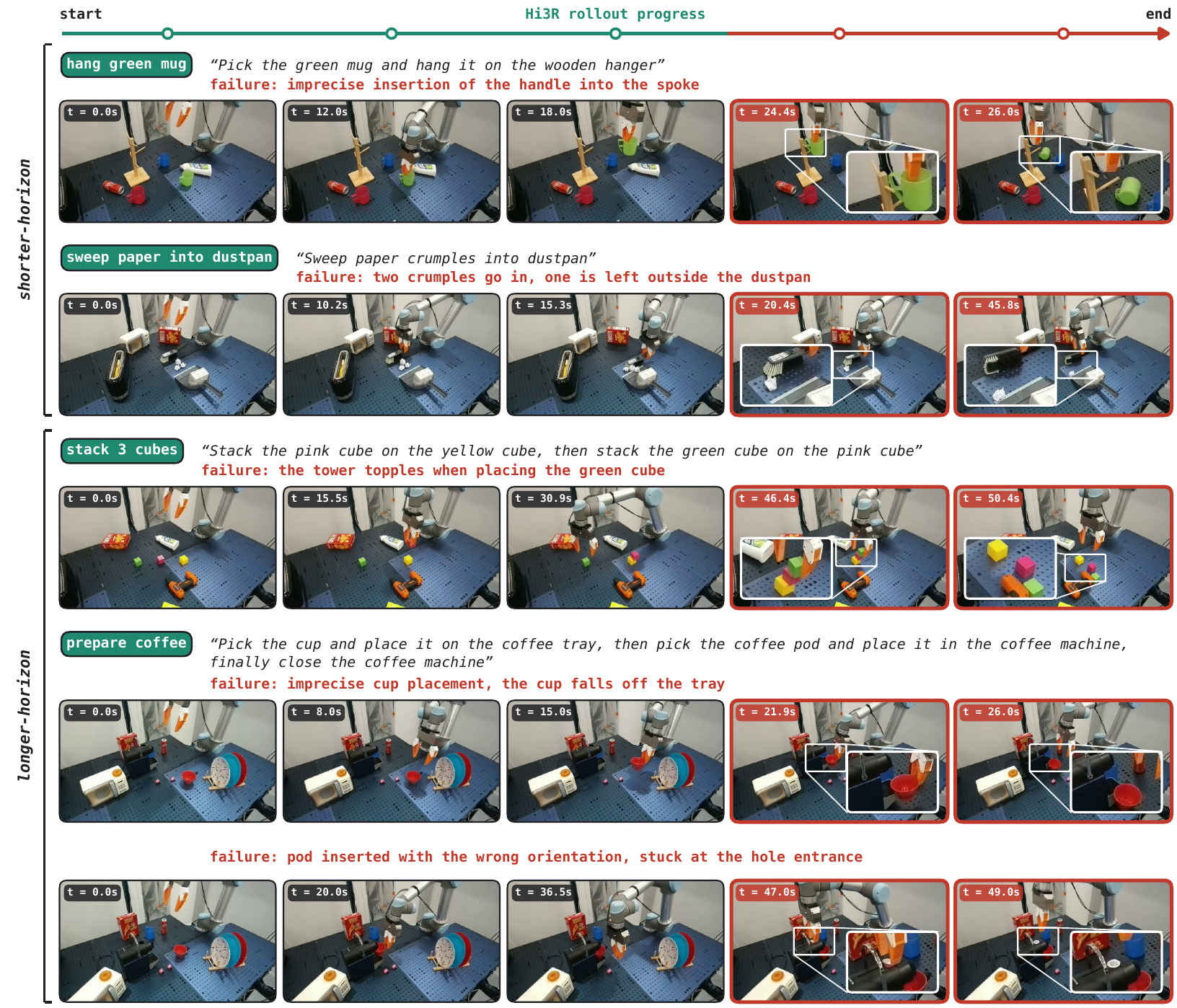}
\caption{\textbf{Qualitative failure examples.} Failed rollouts of Hi3R on our UR5 robotic arm across four
manipulation tasks.}
\label{fig:UR5_failed_rollouts}
\end{figure*}

\subsection{Data Collection and Training}
\label{app:realworld_training}

\textbf{Data Collection.} 
We design 5 tasks as shown in \Cref{app:realworld_eval_protocol}. 
For each task, a human teleoperator collects 50 expert demonstrations using a Meta Quest mask along with its hand controller. Training demonstrations are post-processed to remove idle frames. 

\noindent \textbf{Training.}  
We use forward kinematics of the UR5 arm to compute the end-effector pose in the robot base frame and transform the observed point cloud into the same coordinate frame.
The computed end-effector pose serves as the auxiliary spatial anchor in \ourmodel.
We fine-tune \ourmodel from the DROID-pretrained checkpoint on all collected demonstrations for 10 epochs (approximately 20k iterations), with a batch size of 128.
All other training configurations follow~\Cref{tab:training_config}.

\subsection{Evaluation Protocol and Results}
\label{app:realworld_eval_protocol}

We manually measure the full task success rate and partial credit for each task.
Partial credit is awarded for each completed subtask, such as a successful grasp or placement.
An episode counts as a full success only when every subtask is completed in the correct order.

\begin{itemize}
    \item \textbf{place\_toasts\_in\_toaster}: place the two toasts in the toaster, then turn on the toaster.
    Partial credit: 0.2 for grasp; 0.2 for placement; 0.2 for turning on the toaster.
    \item \textbf{stack\_3\_cubes}: stack the pink cube on the yellow cube, then the green cube on the pink cube.
    Partial credit: 0.25 per grasp and 0.25 per placement for each of the two cubes; $-0.25$ for wrong order.
    \item \textbf{hang\_green\_mug\_onto\_hanger}: pick the green mug and hang it on the wooden hanger.
    Partial credit: 0.5 for grasp; 0.5 for placement.
    \item \textbf{sweep\_paper\_into\_dustpan}: grasp the brush and sweep the paper crumples into the dustpan.
    Partial credit: 0.5 for grasp; 0.25 once one crumple is inside the dustpan; 0.25 once all crumples are inside.
    \item \textbf{prepare\_coffee}: pick the cup and place it on the coffee tray, then pick the coffee pod and place it in the coffee machine, finally close the coffee machine.
    Partial credit: 0.2 per grasp; 0.2 per placement; 0.1 for precisely aligning the pod with the hole; 0.1 for closing the machine.
\end{itemize}

\section{Qualitative Results}
\label{app:qualitative}

We present successful and failed real-robot rollouts of \ourmodel on the UR5 in
\Cref{fig:UR5_successful_rollouts} and \Cref{fig:UR5_failed_rollouts}, respectively.
More examples are in the supplementary video.

\begin{figure*}
    \centering
    \includegraphics[width=1\linewidth]{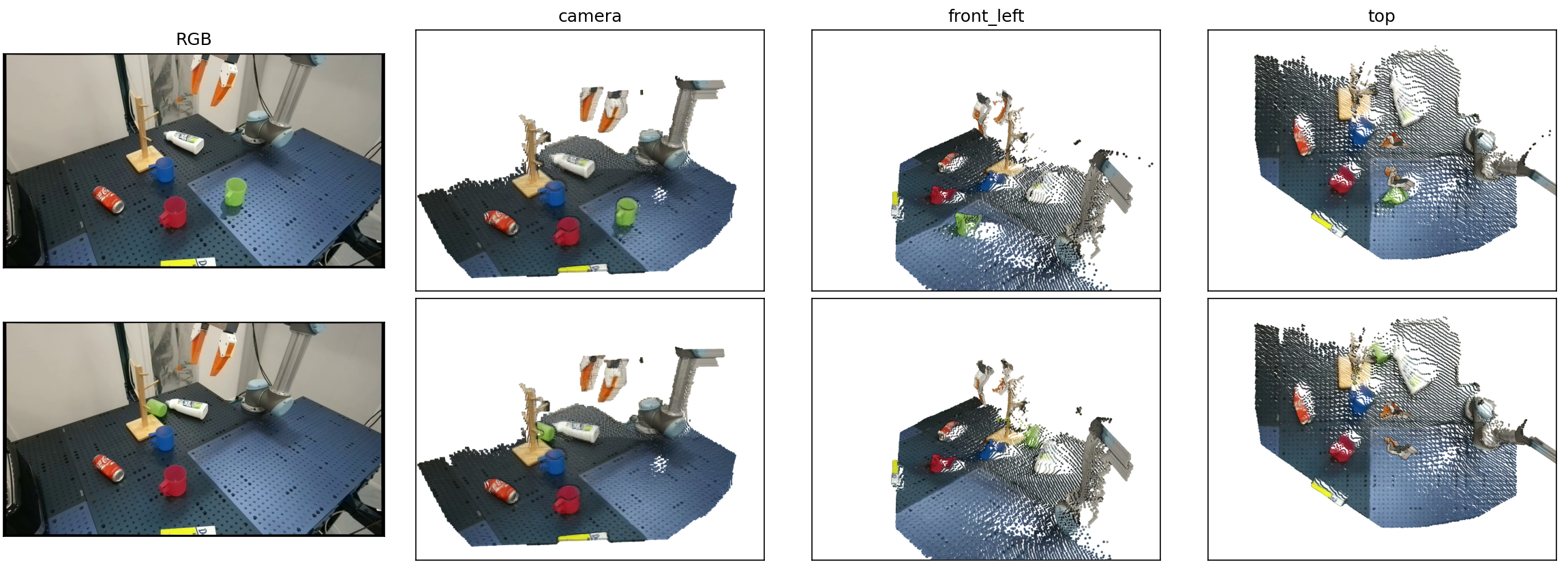}
    \caption{Noisy point clouds for the hang green mug onto hanger task in the real world.}
    \label{fig:noisy_point_clouds}
\end{figure*}

We observe three main challenges underlying the failures:
(1) \textit{Noisy and incomplete point clouds.}
Depth noise and partial observations can lead to inaccurate scene geometry, as illustrated by the noisy mug point cloud in \Cref{fig:noisy_point_clouds}.
More reliable depth sensing and multi-view observations could mitigate this issue.
(2) \textit{Limited failure recovery.}
For the sweeping task, the demonstrations contain only single-attempt executions.
As a result, when the robot fails to sweep all paper crumples in one attempt, it struggles to perform corrective actions.
Training on recovery demonstrations or using reinforcement learning could improve robustness to such failures.
(3) \textit{Object orientation understanding.}
While \ourmodel improves spatial precision, it does not explicitly model object orientation, which can lead to incorrect grasping or manipulation poses as shown in the prepare coffee failed example.
Training on more diverse data with greater variation in object poses could improve orientation understanding and manipulation robustness.

\section{Limitations and Future Work}
\label{app:limitation}

Our approach relies on calibrated RGB-D cameras, and thus noisy depth estimates and calibration may affect performance.
Due to limited computational resources, we only pretrain on a subset of DROID and adopt a relatively lightweight training setup. Scaling up pretraining data, training configuration, and model size could further improve performance, given the promising results achieved with our current setup.

Beyond scaling, future directions include integrating our policy into agentic systems for long-horizon manipulation, improving control through reinforcement learning, extending the shared 3D representation to additional modalities such as tactile sensing, and incorporating world modeling to predict future scene dynamics.

\end{document}